\documentclass[10pt,twocolumn,letterpaper]{article}

\usepackage{cvpr}
\usepackage{times}
\usepackage{epsfig}
\usepackage{graphicx}
\usepackage{amsmath}
\usepackage{amssymb}

\usepackage{url}
\usepackage{xspace}
\usepackage{booktabs}
\usepackage{setspace}
\usepackage{tabularx}
\usepackage[noend]{algpseudocode}
\usepackage{algorithm,algorithmicx}
\usepackage{multirow}
\usepackage{balance}
\usepackage{setspace}

\usepackage[pagebackref=true,breaklinks=true,letterpaper=true,colorlinks,bookmarks=false,urlcolor=black]{hyperref}

\cvprfinalcopy 

\def\cvprPaperID{186} 

\def\frame{\mathcal{X}}
\def\intent{{\mathcal{X}\!\mathcal{I}}}
\def\qintent{{\mathcal{I}}}
\def\pair{{\mathcal{X}\!\mathcal{X}}}

\newcommand{\myparagraph}[1]{\vspace{.3em}\noindent{\bf #1}}

\newcommand{\myurl}[1]{\texttt{#1}}

\ifcvprfinal\fi

\begin{document}

\title{Asynchronous Temporal Fields for Action Recognition}

\renewcommand{\thefootnote}{\fnsymbol{footnote}}
\author{Gunnar A. Sigurdsson$^1\footnotemark[1]$ \ \ \ \ 
Santosh Divvala$^{2,3}$ \ \ \ 
Ali Farhadi$^{2,3}$ \ \ \ 
Abhinav Gupta$^{1,3}$ \\
$^1$Carnegie Mellon University \ 
$^2$University of Washington \
$^3$Allen Institute for Artificial Intelligence \\
\myurl{github.com/gsig/temporal-fields/}
}
\thispagestyle{empty}
\maketitle
\thispagestyle{empty}
\begin{abstract}
Actions are more than just movements and trajectories: we cook to eat and we hold a cup to drink from it. A thorough understanding of videos requires going beyond appearance modeling and necessitates reasoning about the sequence of activities, as well as the higher-level constructs such as intentions. But how do we model and reason about these? We propose a fully-connected temporal CRF model for reasoning over various aspects of activities that includes objects, actions, and intentions, where the potentials are predicted by a deep network. 
End-to-end training of such structured models is a challenging endeavor: For inference and learning we need to construct mini-batches consisting of whole videos, leading to mini-batches with only a few videos. This causes high-correlation between data points leading to breakdown of the backprop algorithm.  
To address this challenge, we present an asynchronous variational inference method that allows efficient end-to-end training. Our method achieves a classification mAP of 22.4\% on the {\em Charades}~\cite{charades} benchmark, outperforming the state-of-the-art (17.2\% mAP), and offers equal gains on the task of temporal localization.
\end{abstract} 

\section{Introduction} 
\label{sec:intro}
\footnotetext{\footnotemark[1]Work was done while Gunnar was an intern at AI2.}

\begin{figure}[t]
\includegraphics[width=\linewidth]{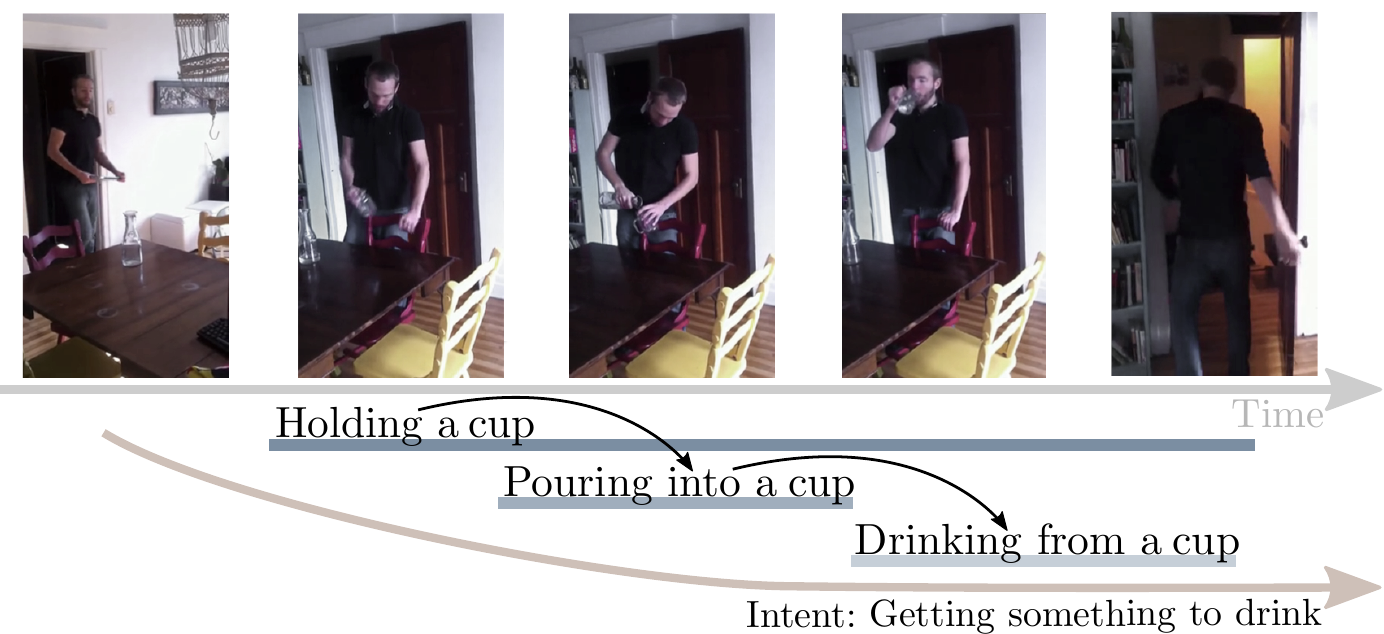}
\caption{Understanding human activities in videos requires jointly reasoning about multiple aspects of activities, such as `what is happening', `how', and `why'. In this paper, we present an end-to-end deep structured model over time trained in a stochastic fashion. The model captures rich semantic aspects of activities, including \emph{Intent} (why), \emph{Category} (what), \emph{Object} (how). The figure shows video frames and annotations used in training from the {\em Charades}~\cite{charades} dataset. }
\label{fig:teaser1}
\end{figure}

Consider the video shown in Figure~\ref{fig:teaser1}: A man walks through a doorway, stands at a table, holds a cup, pours something into it, drinks it, puts the cup on the table, and finally walks away. Despite depicting a simple activity, the video involves a rich interplay of a sequence of actions with underlying goals and intentions. For example, the man stands at the table `to take a cup', he holds the cup `to drink from it', etc. Thorough understanding of videos requires us to model such interplay between activities as well as to reason over extensive time scales and multiple aspects of actions (objects, scenes, etc). 

Most contemporary deep learning based methods have treated the problem of video understanding as that of only appearance and motion (trajectory) modeling~\cite{Simonyan13,Tran15,Georgia15,Snoek16}. While this has fostered interesting progress in this domain, these methods still struggle to outperform models based on hand-crafted features, such as Dense Trajectories~\cite{WangIDT13}. 
Why such a disconnect? We argue that video understanding requires going beyond appearance modeling, and necessitates reasoning about the activity sequence as well as higher-level constructs such as intentions. The recent emergence of large-scale datasets containing rich sequences of realistic activities~\cite{charades,yeung2015every,weinzaepfel2016towards} comes at a perfect time facilitating us to explore such complex reasoning. 

But what is the right way to model and reason about temporal relations and goal-driven behaviour? Over the last couple of decades, graphical models such as Conditional Random Fields (CRFs) have been the prime vehicles for structured reasoning. 
Therefore, one possible alternative is to use ConvNet-based approaches~\cite{alexnet12} to provide features for a CRF training algorithm. 
Alternatively, it has been shown that integrating CRFs with ConvNet architectures and training them in an end-to-end manner provides substantial improvements in tasks such as segmentation and situation recognition~\cite{raqueldense2016,ChenSchwingICML2015,yatskarsituation}. 

Inspired by these advances, we present a deep-structured model that can reason temporally about multiple aspects of activities. For each frame, our model infers the activity category, object, action, progress, and scene using a CRF, where the potentials are predicted by a jointly end-to-end trained ConvNet over all predictions in all frames. This CRF has a latent node for the intent of the actor in the video and pair-wise relationships between all individual frame predictions. 

While our model is intuitive, training it in an end-to-end manner is a non-trivial task. Particularly, end-to-end learning requires computing likelihoods for individual frames and doing joint inference about all connected frames with a CRF training algorithm. 
This is in stark contrast with the standard stochastic gradient descent (SGD) training algorithm (backprop) for deep networks, where we require mini-batches with a large number of independent and uncorrelated samples, not just a few whole videos.
In order to handle this effectively: (1) we relax the Markov assumption and choose a fully-connected temporal model, such that each frame's prediction is influenced by all other frames, and (2) we propose an asynchronous method for training fully-connected structured models for videos. Specifically, this structure allows for an implementation where the influence (messages) from other frames are approximated by emphasizing influence from frames computed in recent iterations. They are more accurate, and show advantage over being limited to only neighboring frames. In addition to being more suitable for stochastic training, fully-connected models have shown increased performance on various tasks~\cite{densecrfsegmentation,raqueldense2016}.

In summary, our key contributions are: (a) a deep CRF based model for structured understanding and comprehensive reasoning of videos in terms of multiple aspects, such as action sequences, objects, and even intentions; (b) an asynchronous training framework for expressive temporal CRFs that is suitable for end-to-end training of deep networks; and, (c) substantial improvements over state-of-the-art, increasing performance from 17.2\% mAP to 22.4\% mAP on the challenging Charades~\cite{charades} benchmark.

\section{Related Work} 

Understanding activities and actions has an extensive history~\cite{poppe2010survey,weinland2011survey,STIP05,HOG3D,HOF,MBH06,Matikainen09,WangIDT13,Peng14,Lan15}. Interestingly, 
analyzing actions by their appearance has gone through multiple iterations. Early success was with hand-crafted representations such as Space Time Interest Points (STIP)~\cite{STIP05}, 3D Histogram of Gradient (HOG3D)~\cite{HOG3D}, Histogram of Optical Flow (HOF)~\cite{HOF}, and Motion Boundary Histogram~\cite{MBH06}. These methods capture and analyze local properties of the visual-temporal datastream. In the past years, the most prominent hand-crafted representations have been from the family of trajectory based approaches~\cite{Matikainen09,WangIDT13,Peng14,Lan15}, where the Improved Dense Trajectories (IDT)~\cite{WangIDT13} representation is in fact on par with state-of-the-art on multiple recent datasets~\cite{THUMOS15,charades}. 

Recently there has been a push towards mid-level representations of video~\cite{Corso12,YaleS13,ArpitJ13,lan2015iccv}, that capture beyond local properties. However, these approaches still used hand-crafted features. With the advent of deep learning, learning representations from data has been extensively studied~\cite{3DCNN,Karpathy14,2stream14,TDD15,Taylor10,Tran15,Le11,Georgia15,Xu15,Vondrick16_repr,scnn_shou_wang_chang_cvpr16,Souza16}. Of these, one of the most popular frameworks has been the approach of Simonyan et al.~\cite{2stream14}, who introduced the idea of training separate color and optical flow networks to capture local properties of the video. 

Many of those approaches were designed for short clips of individual activities and hence do not generalize well to realistic sequences of activities. 
Capturing the whole information of the video in terms of temporal evolution of the video stream has been the focus of some recent approaches~\cite{KevinT12,Basura15,Izadinia12,Rohrbach12,Sun13,pirsiavash2014parsing}.
Moving towards more expressive deep networks such as LSTM has become a popular method for encoding such temporal information~\cite{Srivastava15,Donahue15,cnnlstm,Sun_2015_ICCV,Wang_Transformation,sigurdsson2016learning,yeung2015end}. Interestingly, while those models move towards more complete understanding of the full video stream, they have yet to significantly outperform local methods~\cite{2stream14} on standard benchmarks.

A different direction in understanding comes from reasoning about the complete video stream in a complementary direction --- Structure. Understanding activities in a human-centric fashion encodes our particular experiences with the visual world. Understanding activities with emphasis on objects has been a particularly fruitful direction~\cite{li2007and,ryoo2007hierarchical,gupta2009observing,prest2012weakly,Vondrick16_actns}. In a similar vein, some works have also tried modeling activities as transformations~\cite{Wang_Transformation} or state changes~\cite{Fathi13}. Recently, there has been significant progress in modelling the complete human-centric aspect, where image recognition is phrased in terms of objects and their roles~\cite{yatskarsituation,VSRL_gupta15}. Moving beyond appearance and reasoning about the state of {\em agents} in the images requires understanding human intentions~\cite{kitani2012activity,pirsiavash2014inferring}. This ability to understand people in terms of beliefs and intents has been traditionally studied in psychology as the Theory of mind~\cite{premack1978does}.

How to exactly model structure of the visual and temporal world has been the pursuit of numerous fields. Of particular interest is work that combines the representative power of deep networks with structured modelling. Training such models is often cumbersome due to the differences in jointly training deep networks (stochastic sampling) and sequential models (consecutive samples)~\cite{mnih2013playing,raqueldense2016}.
In this work, we focus on fully-connected random fields, that have been popular in image segmentation~\cite{densecrfsegmentation}, where image filtering was used for efficient message passing, and later extended to use CNN potentials~\cite{schwing2015fully}.

\section{Proposed Method} 
\label{sect:app}

\begin{figure*}[t]
\centering
\includegraphics[width=1.0\linewidth]{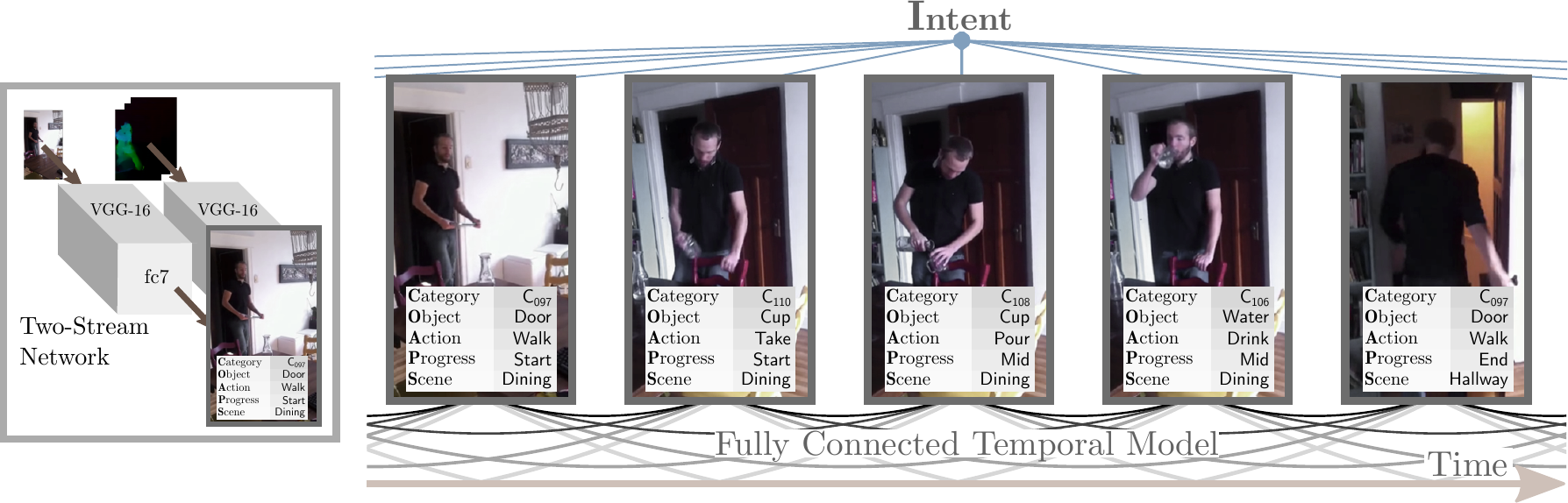}
\caption{An overview of our structured model. The semantic part captures \emph{object}, \emph{action}, etc. at each frame, and temporal aspects captures those over time. On the left side, we show how for each timepoint in the video, a Two-Stream Network predicts the potentials. Our model jointly reasons about multiple aspects of activities in all video frames. The \emph{Intent} captures groups of activities of the person throughout the whole sequence of activities, and fine-grained temporal reasoning is through fully-connected temporal connections.}
\label{fig:model}
\end{figure*}

Given a video with multiple activities, our goal is to understand the video in terms of activities. Understanding activities requires reasoning about objects being interacted with, the place where the interaction is happening, what happened before and what happens after this current action and even the intent of the actor in the video. We incorporate all these by formulating a deep Conditional Random Field (CRF) over different aspects of the activity over time. That is, a video can be interpreted as a graphical model, where the components of the activity in each frame are nodes in the graph, and the model potentials are the edges in the graph. 

In particular, we create a CRF which predicts activity, object, etc., for every frame in the video. For reasoning about time, we create a \emph{fully-connected temporal CRF}, referred as Asynchronous Temporal Field in the text. That is, unlike a linear-chain CRF for temporal modelling (the discriminative counterpart to Hidden Markov Models), each node depends on the state of every other node in the graph. We incorporate intention as another latent variable which is connected to all the action nodes. 
This is an unobserved variable that influences the sequence of activities. This variable is the common underlying factor that guides and better explains the sequence of actions an agent takes. Analysis of what structure this latent variable learns is presented in the experiments.
Our model has three advantages: (1) it addresses the problem of long-term interactions; (2) it incorporates reasoning about multiple parts of the activity, such as objects and intent; and (3) more interestingly, as we will see, it allows for efficient end-to-end training in an asynchronous stochastic fashion.

\subsection{Architecture}
In this work we encode multiple components of an activity. Each video with $T$ frames is represented as $\{X_1, \dots, X_T, I\}$ where $X_t$ is a set of frame-level random variables for time step $t$ and $I$ is an unobserved random variable that represent global intent in the entire video. We can further write $X_t = \{C_t,O_t,A_t,P_t,S_t\}$, where $C$ is the activity category (e.g., `drinking from cup'), $O$ corresponds to the object (e.g., `cup'), $A$ represents the action (e.g., `drink'), $P$ represents the progress of the activity \{start, middle, end\}, and $S$ represents the scene (e.g. `Dining Room'). 
For clarity in the following derivation we will refer to all the associated variables of $X_t$ as a single random variable $X_t$. A more detailed description of the CRF is presented in the appendix.

Mathematically we consider a random field $\{X,I\}$ over all the random variables in our model ($\{X_1, \dots, X_T, I\}$). Given an input video $V{=}\{V_1,\dots,V_T\}$, where $V_t$ is a video frame, our goal is to estimate the maximum a posteriori labeling of the random field by marginalizing over the intent $I$. This can be written as:
\begin{equation}
\mathbf{x}^* = \mathrm{arg}\,\max_x \sum_I P(x,I|V).
\end{equation}
For clarity in notation, we will drop the conditioning on $V$ and write $P(X,I)$. We can define $P(X,I)$ using Gibbs distribution as: $P(X,I) {=} \frac{1}{Z(\mathbf{V})} \exp \left( -E(x,I) \right)$ where $E(x,I)$ is the Gibbs energy over $x$. In our CRF, we model all unary and pairwise cliques between all frames $\{X_1,\dots,X_T\}$ and the intent $I$. The Gibbs energy is:
\begingroup\makeatletter\def\f@size{8}\check@mathfonts
\def\maketag@@@#1{\hbox{\m@th\normalsize\normalfont#1}}%
\begin{align}
E(\mathbf{x},I) = \underbrace{\sum_i \phi_\frame(x_i)}_{\mathrm{Semantic}} + \underbrace{\sum_{i} \phi_\intent(x_i,I) + \sum_{\substack{i,j\\i \neq j}} \phi_\pair(x_i,x_j)}_{\mathrm{Temporal}},
\end{align}
\endgroup
where $\phi_\pair(x_i,x_j)$ is the potential between frame $i$ and frame $j$, and $\phi_\intent(x_i,I)$ is the potential between frame $i$ and the intent. For notational clarity $\phi_\frame(x_i)$ incorporates all unary and pairwise potentials for ${C_t,O_t,A_t,P_t,S_t}$. 
The model is best understood in terms of two aspects: Semantic aspect, which incorporates the local variables in each frame (${C_t,O_t,A_t,P_t,S_t}$); and Temporal aspect, which incorporates interactions among frames and the intent $I$. This is visualized in Figure~\ref{fig:model}. 
We will now explain the semantic, and temporal potentials.

\myparagraph{Semantic aspect}
The frame potential $\phi_\frame(x_i)$ incorporates the interplay between activity category, object, action, progress and scene, and could be written explicitly as $\phi_\frame(C_t,O_t,A_t,P_t,S_t)$. In practice this potential is composed of unary, pairwise, and tertiary potentials directly predicted by a CNN. We found predicting only the following terms to be sufficient without introducing too many additional parameters:
$\phi_\frame(C_t,O_t,A_t,P_t,S_t){=}\phi(O_t,P_t){+}\phi(A_t,P_t){+}\phi(O_t,S_t)+ \phi(C_t,O_t,A_t,P_t)$ where we only model the assignments seen in the training set, and assume others are not possible.

\begin{figure}
\centering
\includegraphics[width=1.0\linewidth]{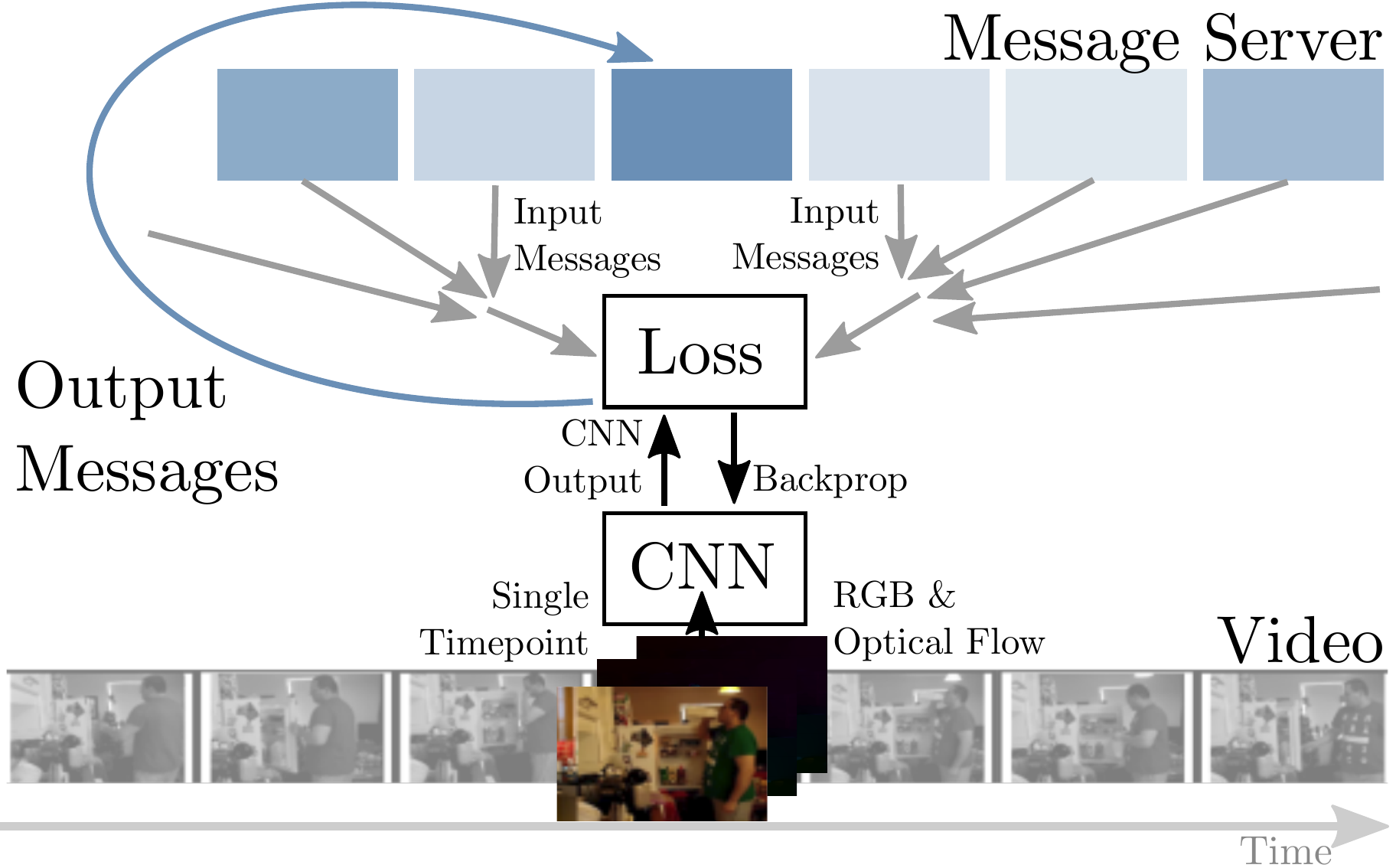}
\caption{Illustration of the learning algorithm, and the message passing structure. Each timepoint that has been processed has a message (Blue highlights messages that have recently been computed). The loss receives a combination of those messages, uses those to construct new messages, and updates the network. }
\label{fig:learning}
\end{figure}

\myparagraph{Temporal aspect}
The temporal aspect of the model is both in terms of the frame-intent potentials $\phi_{\intent}(x_i,I)$ and frame-frame potentials $\phi_{\pair}(x_i,x_j)$. The frame-intent potentials are predicted with a CNN from video frames (pixels and motion). The pairwise potentials $\phi_{\pair}(x_i,x_j)$ for two time points $i$ and $j$ in our model have the form:
\begin{align}
\phi_{\pair}(x_i,x_j) = \mu(x_i,x_j)\sum_{m} w^{(m)}k^{(m)}(v_i,v_j),
\end{align}
where $\mu$ models the asymmetric affinity between frames, $w$ are kernel weights, and each $k^{(m)}$ is a Gaussian kernel that depends on the videoframes $v_i$ and $v_j$. In this work we use a single kernel that prioritises short-term interactions:
\begin{align}
\label{eq:sigma}
k(v_i,v_j) = \exp \left( - \frac{(j-i)^2}{2\sigma^2}\right)
\end{align}
The parameters of the general asymmetric compatibility function $\mu(x_i,x_j)$ are learned from the data, and $\sigma$ is a hyper-parameter chosen by cross-validation.

\subsection{Inference}

While it is possible to enumerate all variable configurations in a single frame, doing so for multiple frames and their interactions is intractable. Our algorithm uses a structured variational approximation to approximate the full probability distribution. In particular, we use a mean-field approximation to make inference and learning tractable. With this approximation, we can do inference by keeping track of message between frames, and asynchronously train one frame at a time (in a mini-batch fashion). 

More formally, instead of computing the exact distribution $P(X,I)$ presented above, the structured variational approximation finds the distribution $Q(X,I)$ among a given family of distributions that best fits the exact distribution in terms of KL-divergence. By choosing a family of tractable distributions, it is possible to make inference involving the ideal distribution tractable. Here we use $Q(X,I)=Q_\qintent(I) \prod_i Q_i(x_i)$, the structured mean-field approximation. Minimizing the KL-divergence between those two distributions yields the following iterative update equation:
\begingroup\makeatletter\def\f@size{8}\check@mathfonts
\def\maketag@@@#1{\hbox{\m@th\normalsize\normalfont#1}}%
\begin{align}
Q_i(x_i) \propto \exp \bigg\{ & \phi_\frame(x_i) + \mathrm{E}_{U\sim Q_\qintent} \left[ \phi_\intent(x_i,U) \right] \nonumber \\ 
&+ \sum_{j > i} \mathrm{E}_{U_j\sim Q_j} \left[ \phi_{\pair}(x_i,U_j) \right] \bigg\} \nonumber \\
&+ \sum_{j < i} \mathrm{E}_{U_j\sim Q_j} \left[ \phi_{\pair}(U_j,x_i) \right] \bigg\} \\
Q_\qintent(I) \propto \exp \bigg\{ & \sum_j \mathrm{E}_{U_j\sim Q_j} \left[ \phi_\intent(U_j,I) \right] \bigg\}
\label{eq:Q}
\end{align}
\endgroup
where $Q_i$ is marginal distribution with respect to each of the frames, and $Q_\qintent$ is the marginal with respect to the intent. An algorithmic implementation of this equation is as presented in Algorithm~\ref{alg:inference}.

\begin{algorithm}
  \caption{Inference for Asynchronous Temporal Fields
    \label{alg:inference}}
    {\footnotesize
  \begin{algorithmic}[1]
      \State Initialize Q \Comment{Uniform distribution}
      \While{$\textrm{not converged}$}
        \State Visit frame $i$
        \State Get $\sum_{j > i} \mathrm{E}_{U_j\sim Q_j} \left[ \phi_\pair (x_i,U_j) \right]$ 
        \State Get $\sum_{j < i} \mathrm{E}_{U_j\sim Q_j} \left[ \phi_\pair (U_j,x_i) \right]$ 
        \State Get $\sum_{j} \mathrm{E}_{U_j \sim Q_j} \left[ \phi_\intent (U_j,I) \right]$ 
        \While{$\textrm{not converged}$}
          \State Update $Q_i$ and $Q_\qintent$ using Eq.~\ref{eq:Q}
        \EndWhile
        \State Send $\mathrm{E}_{U \sim Q_i} \left[ \phi_\pair (x,U) \right]$
        \State Send $\mathrm{E}_{U \sim Q_i} \left[ \phi_\pair (U,x) \right]$
        \State Send $\mathrm{E}_{U \sim Q_i} \left[ \phi_\intent (U,I) \right]$
      \EndWhile
  \end{algorithmic}
  }
\end{algorithm}

\noindent Here `Get' and `Send' refer to the message server, and $f(x)$ is a message used later by frames in the same video. The term message server is used for a central process that keeps track of what node in what video sent what message, and distributes them accordingly when requested. In practice, this could be implemented in a multi-machine setup.

\begin{figure}
\centering
\includegraphics[width=1.0\linewidth]{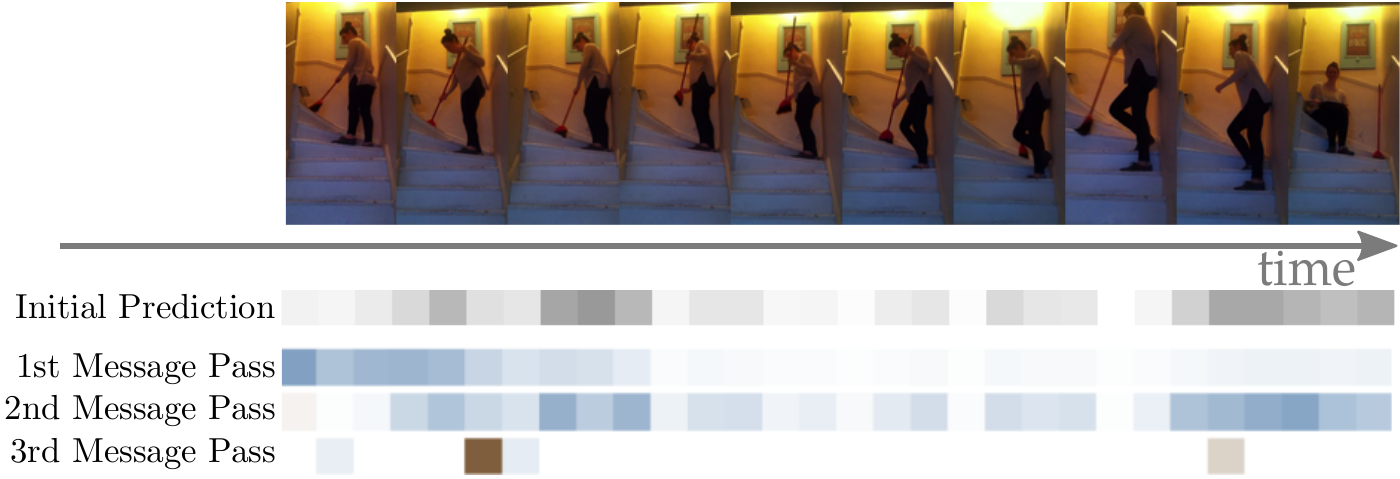}
\caption{Evolution of prediction with increasing messages passes. The first row shows the initial prediction for the category {\em tidying with a broom} without any message passing, where darker colors correspond to higher likelihood, blue is then an increase in likelihood, and brown decrease. In the first message pass, the confidence of high predictions gets spread around, and eventually increases the confidence of the whole prediction.}
\label{fig:messages}
\end{figure}

\subsection{Learning}

Training a deep CRF model requires calculating derivatives of the objective in terms of each of the potentials in the model, which in turn requires inference of $P(X,I|V)$. The network is trained to maximize the log-likelihood of the data $l(X) = \log \sum_I P(x,I|V)$. 
The goal is to update the parameters of the model, for which we need gradients with respect to the parameters. Similar to SGD, we find the gradient with respect to one part of the parameters at a time, specifically with respect to one potential in one frame. That is, $\phi_\frame^i(\hat{x})$ instead of $\phi_\frame(\hat{x})$.
The partial derivatives of this loss with respect to each of the potentials are as follows:
\begingroup\makeatletter\def\f@size{8}\check@mathfonts
\def\maketag@@@#1{\hbox{\m@th\normalsize\normalfont#1}}%
\begin{align}
\frac{\partial l(X)}{\partial \phi_\frame^i(\hat{x})} &=
\mathbf{1}_{x=\hat{x}} - Q_i(\hat{x}) \label{eq:gradients1}\\ 
\frac{\partial l(X)}{\partial \phi_\intent^i(\hat{x},\hat{I})} &= \frac{\exp \sum_j \phi_\intent(x_j,\hat{I})}{\sum_I \exp \sum_j  \phi_\intent(x_j,I)}\mathbf{1}_{x=\hat{x}} - Q_i(\hat{x}) Q_\qintent(\hat{I}) \label{eq:gradients2}\\
\frac{\partial l(X)}{\partial \mu^i(a,b)} &= 
\sum_{j>i} \mathbf{1}_{x=a} k(v_i,v_j) - Q_i(\hat{x}) \sum_{j>i} Q_\qintent(b) k(v_i,v_j) \nonumber \\
&+ \sum_{j<i} \mathbf{1}_{x=b} k(v_j,v_i) - Q_i(\hat{x}) \sum_{j<i} Q_\qintent(a) k(v_i,v_j)
\label{eq:gradients3}
\end{align}
\endgroup
where $\phi_\frame^i(\hat{x})$ and $\phi_\intent^i(\hat{x},\hat{I})$ is the frame and frame-intent potentials of frame $i$, and we use $\hat{x}$ to distinguish between the labels and variables the derivative is taken with respect to. $\mu^i(a,b)$ are the parameters of the asymmetric affinity kernel with respect to frame $i$, and $\mathbf{1}_{x=\hat{x}}$ is a indicator variable that has the value one if the ground truth label corresponds to the variable. Complete derivation is presented in the appendix. These gradients are used to update the underlying CNN model. These update equations lead to the learning procedure presented in Algorithm~\ref{alg:learning}.
\algnewcommand\algorithmicforeach{\textbf{for each}}
\algdef{S}[FOR]{ForEach}[1]{\algorithmicforeach\ #1\ \algorithmicdo}
\begin{algorithm}
  \caption{Learning for Asynchronous Temporal Fields
    \label{alg:learning}}
    {\footnotesize
  \begin{algorithmic}[1]
      \State Given videos $\mathcal{V}$ 
      \While{$\textrm{not converged}$}
        \ForEach{example in mini-batch}
          \State Sample frame $v \in \mathbf{V}\subseteq \mathcal{V}$
          \State Get incoming messages
          \State Update $Q_i$ and $Q_\qintent$
          \State Find gradients with Eq.~\ref{eq:gradients1}-\ref{eq:gradients3}
          \State Backprop gradients through CNN
          \State Send outgoing messages
        \EndFor
      \EndWhile
  \end{algorithmic}
  }
\end{algorithm}

Figure~\ref{fig:learning} graphically illustrates the learning procedure. Since the videos are repeatedly visited throughout the training process, we do not have to run multiple message passes to calculate each partial gradient. This shares ideas with contrastive divergence~\cite{hinton2002training,salakhutdinov2009deep}.
Given a single video at test time, we visualize in Figure~\ref{fig:messages} how the predictions changes as the distribution converges with multiple messages passes.

\myparagraph{Message Passing}
The key thing to note is all the incoming messages are of the form $M(z) {=} \sum_j f_j(z)$ where $f_j$ is some function from node $j$; for e.g., $M(z) = \sum_j \mathrm{E}_{U_j \sim Q_j} \left[ \phi_\intent (U_j,z) \right] = \sum_j f_j(z)$ from Algorithm~\ref{alg:inference}. We use the following approximation during training:
\begin{align}
M(z) {\approx} \frac{h}{\sum_j d^j} \sum_j d^j f_{J(j)}(z),
\end{align}
where $d\in[0,1]$ is a discount factor, $h$ is a hyperparameter, and $J(\cdot)$ is an ordering of the messages in that video based on the iteration in which the message was computed. The messages are a weighted combination of stored messages.

\section{Experimental Results and Analysis} 

We analyzed the efficacy of our model on the challenging tasks of video activity classification and temporal localization. In addition, we investigated the different parts of the model, and will demonstrate how they operate together.

\myparagraph{Dataset}
Recent years have witnessed an emergence of large-scale datasets containing sequences of common daily activities~\cite{charades,yeung2015every,weinzaepfel2016towards}. For our evaluation, we chose the {\em Charades} dataset~\cite{charades}. This dataset is a challenging benchmark containing 9,848 videos across 157 action classes with 66,500 annotated activities, including nouns (objects), verbs (actions), and scenes. A unique feature of this dataset is the presence of complex co-occurrences of realistic human-generated activities making it a perfect test-bed for our analysis. We evaluate video classification using the evaluation criteria and code from~\cite{charades}. Temporal localization is evaluated in terms of per-frame classification using the provided temporal annotations.

\begin{figure}
\centering
\includegraphics[width=0.8\linewidth]{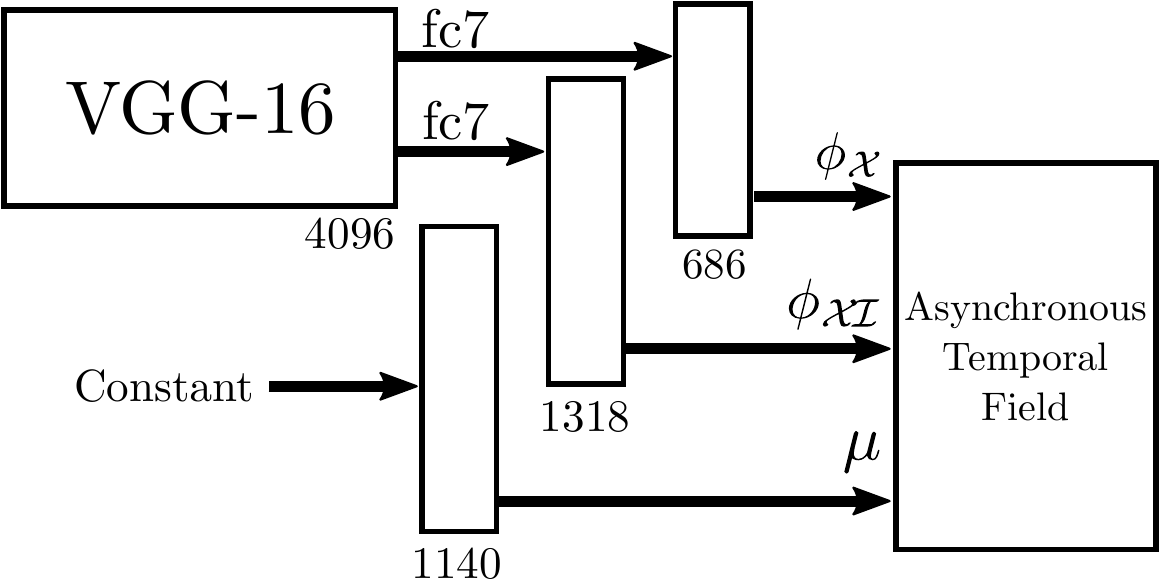}
\caption{The VGG-16 variant predicts the potentials for both RGB and Flow. The network predicts the values of all potentials except one (in this figure we group the frame potentials $\phi_\frame$ into one layer for clarity). The model is trained end-to-end by passing gradients from the Asynchronous Temporal Field through the network.}
\label{fig:CNN}
\end{figure}

\myparagraph{Implementation details}
We use a VGG16 network~\cite{Simonyan15} with additional layers to predict the model potentials (Figure~\ref{fig:CNN}). We train both a network on RGB frames, and stacks of optical flow images, following the two-stream architecture~\cite{2stream14}. The main challenge in training the network is the increase in the output layer size. For the larger potentials, we used the following structure to go from fc7 to $\phi_{\intent}$: Linear layer (4096 to 100), ReLU, Dropout, Linear layer (100 to the potential values). 

The input to the RGB network is an image of size $224{\times}224{\times}3$ where we crop random location, size, and aspect ratio. We use data augmentation with color jitter and PCA lighting noise. The RGB network was pretrained on ImageNet. The input to the Flow network is a stack of 10 consecutive optical flow frames at 24 FPS starting with the current frame. Since each optical flow has two channels, the input size is $224{\times}224{\times}20$ as in~\cite{2stream14}. The Flow network was pretrained on UCF101~\cite{UCF101} as in Sigurdsson et al.~\cite{charades}, and random cropped in the same way as RGB. 

We follow the training setup in Charades~\cite{charades} and consider a frame to have one activity label at a time. Even so, our method is still able to reason about other activities in the video. Convergence of the model is evaluated using the approximate distribution $Q_i(X)$ at each frame. The Charades dataset has the property that scenes were chosen at random for each sequence of activities. For this reason, we found reasoning about scenes to reduce the performance, and the weight of that term was lowered in the model.

To obtain annotations for action progress $p_t$, we split each activity annotation into three equally sized parts. All layers of the network are trained with a batch size of $240$ and a learning rate of $10^{-3}$ (RGB), $10^{-5}$ (Flow). Learning rate was reduced by a factor of 10 every 30k iterations for RGB, and every 140k iterations for Flow. The value of the message decay parameter $d$ was set to $d=0.9$, and the standard deviation $\sigma$ in (\ref{eq:sigma}) was set to $6.25$ sec (150 frames).

For testing, we sampled 25 equally spaced frames from the video and synchronously pass messages between the frames until convergence (10 message passes). The predictions of the RGB and Flow networks are combined in a probabilistic fashion by multiplying their probabilistic predictions for each class. More implementation details may be found in the appendix. The networks were implemented in Torch, and the code is available on project page.

\begin{figure}[t]
\centering
\includegraphics[width=1.0\linewidth]{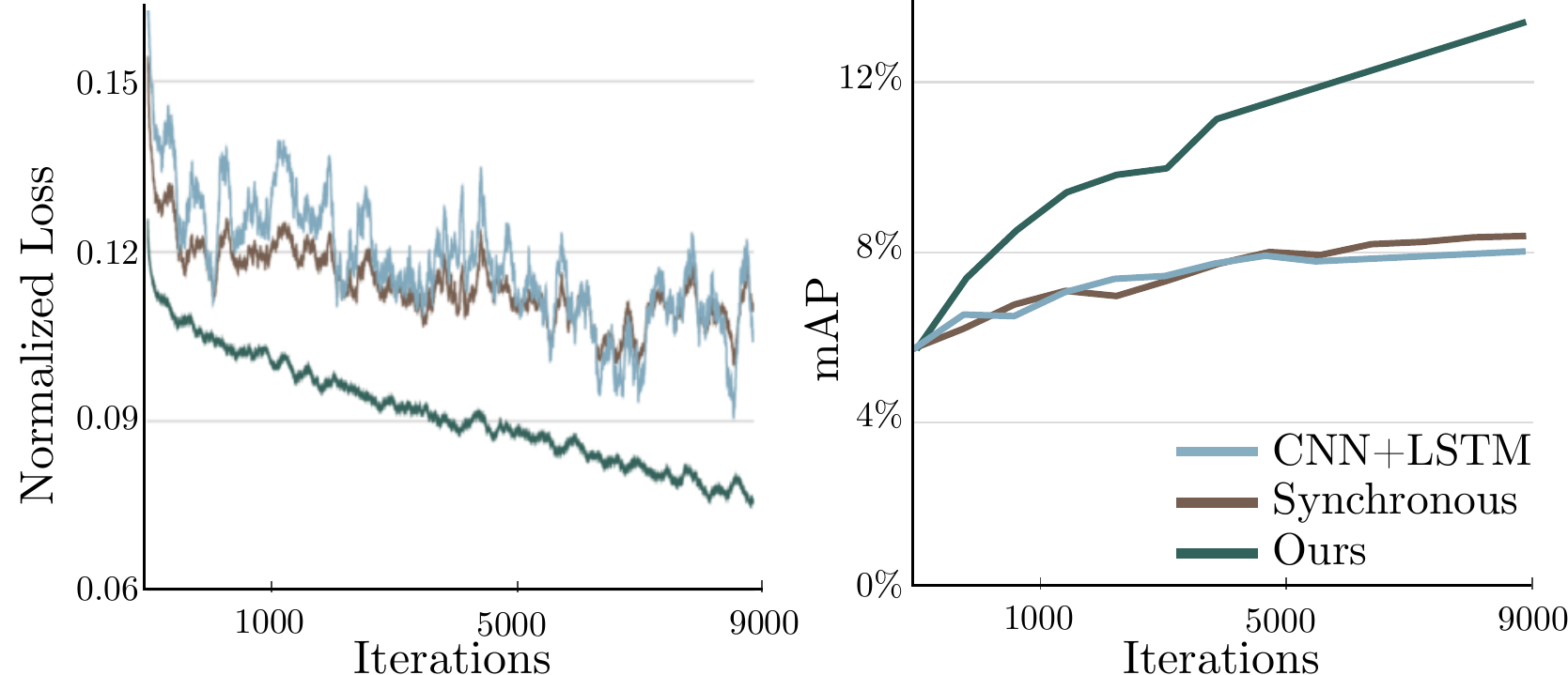}
\caption{Convergence of our method compared to other methods that capture temporal structure. Our asynchronous training method contains more diverse batches, has faster and more stable convergence, and reaches higher accuracy on the test set.}
\label{fig:convergence}
\end{figure}

\myparagraph{Diverse batches}
As highlighted in Section~\ref{sec:intro}, the standard way of sampling batches for temporal models results in high correlation between data points leading to a breakdown of the SGD. To understand the importance of having many diverse examples from multiple videos, we compare the convergence of our method to two alternatives using homogeneous batches: CNN+LSTM from Ng et al.~\cite{cnnlstm}, and a synchronous version of our method, where each batch contains full videos (only three videos fit into each mini-batch). We do synchronous message passing until convergence before calculating gradients for backprop. Figure~\ref{fig:convergence} shows that our asynchronous training method, containing more diverse training batches, has faster and more stable convergence. 

\subsection{Video Classification}

Given a video, the task here is to verify whether it contains one or several of the $157$ activity categories. Classification accuracy is measured with the standard mean average precision (mAP) criterion, where a prediction is given for each video. This task has been shown to be highly challenging, with the state-of-the-art non-ensemble methods reaching an mAP of only $17.2\%$, particularly as each video in this dataset has a sequence of multiple fine-grained activities with a real-world long-tailed activity distribution. 

We trained our models using the provided training split following the procedure outlined in Section~\ref{sect:app}. To make predictions for the whole video, we marginalize out everything except the activity category for 25 equidistant frames in the video. The score for each activity category is the maximum across all frames following the setup from~\cite{charades}. In our analysis, we include the provided non-ensemble baselines from~\cite{charades} as well as the following additional baselines:

{\em Two-Stream++.} We reimplemented the network described in~\cite{charades}, which follows Simonyan et al.~\cite{Simonyan15}, with the same parameters. We added data augmentation and fine-tuned all layers of the network. The performance of only the RGB stream is included ({\em RGB++}). We also consider {\em Two-Stream Extended} which is the same network, but the Flow network was trained for 25 times more iterations than the RGB network (two weeks of computation on a Titan X GPU). Combined with the augmentation, we found this to non-trivially increase the accuracy.

{\em Two-Stream+LSTM.} We followed the method outlined in~\cite{cnnlstm} to jointly train a LSTM on top of the two-stream network. We trained both an RGB and an Optical Flow network using the same setup from~\cite{charades}. The trained networks from Two-Stream++ were used to initialize the models.

\begin{table}[tb]
\begin{center}
{\footnotesize
\begin{tabular}{lcclc} \toprule
Approach & mAP & \hspace{2em} &Approach & mAP \\ \hline \rule{0pt}{3ex}%
Random~\cite{charades} & 5.9 & & RGB++ & 15.6 \\
C3D~\cite{Tran15} & 10.9 & &Two-Stream++ & 16.8\\
AlexNet~\cite{alexnet12} & 11.3 & &Two-Stream+LSTM & 17.8 \\
IDT~\cite{WangIDT13} & 17.2 & &Two-Stream Extended & 18.6\\ 
Two-Stream~\cite{Simonyan13} & 14.3 & & Ours (RGB Only) & 18.3 \\
 & & & Ours & {\bf 22.4}  \\
\bottomrule
\end{tabular}
}
\end{center}
\caption{Video classification results on Charades~\cite{charades}. The left shows the published baselines from~\cite{charades} and the right show additional new baselines. Our proposed approach outperforms all competing methods on this dataset.}
\label{tbl:classification}
\end{table}

\begin{figure}[t]
\centering
\includegraphics[width=1.0\linewidth]{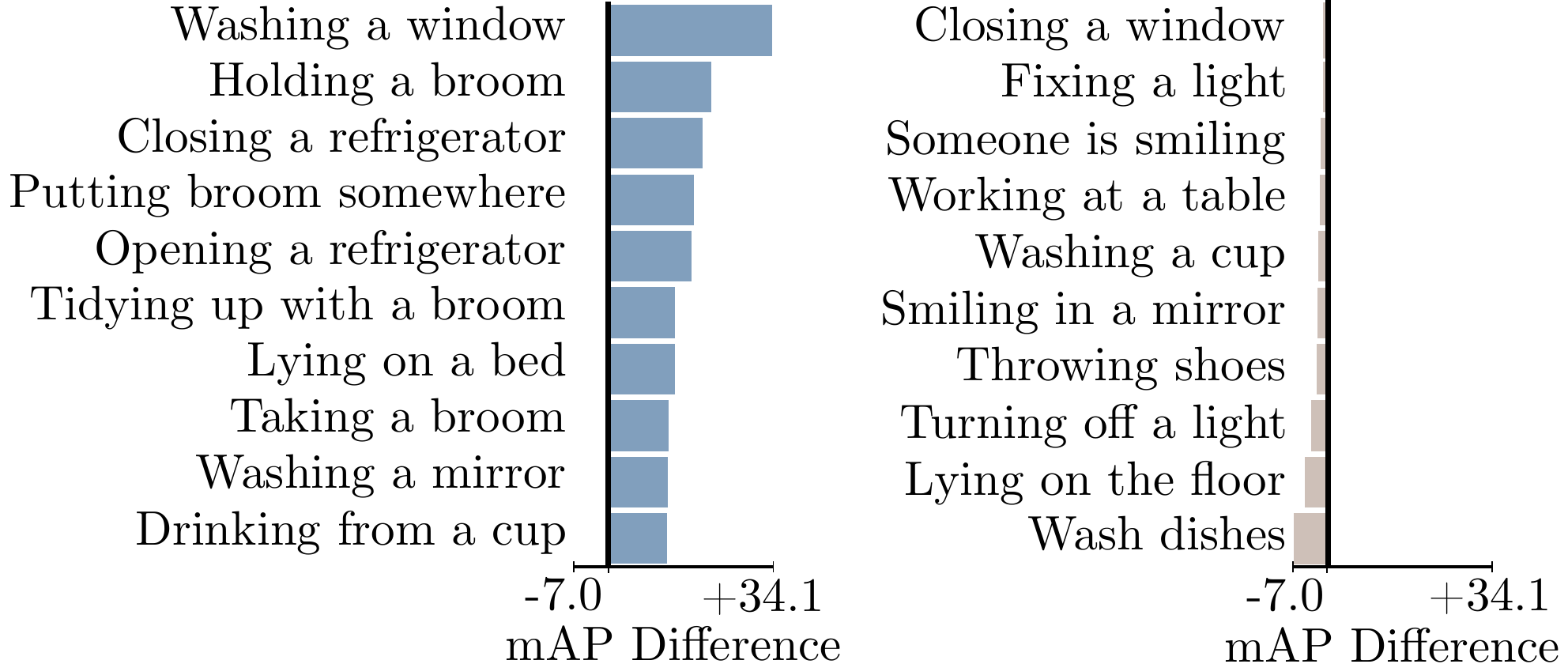}
\caption{The classes with the highest positive and negative difference between our method and Two-Stream (no structure). Our method does better on many classes, without doing much worse on any. In particular, activities that have temporal structure, such as Opening/Closing a refrigerator have significantly higher performance, since our model can reason jointly about those.}
\label{fig:top10bottom10}
\end{figure}

Table~\ref{tbl:classification} displays the accuracy obtained by our method along with the baselines. Our proposed approach obtains an mAP of $22.4\%$ substantially outperforming the Two-stream Extended baseline at $18.6\%$ mAP, and the IDT baseline at $17.2\%$. Our method reasons over significantly larger timescales and multiple aspects of the activities. To ascertain this, we highlight in Figure~\ref{fig:top10bottom10}, the activity classes with the highest positive and negative difference between our method and the Two-Stream network.
It is interesting to note that two of those activities are {\em opening} and {\em closing} a refrigerator, that arguably have a significant causal structure (an {\em open} refrigerator was {\em opened} at some point), which our model harnesses to significantly increase the accuracy.

\begin{figure}[tb]
\centering
\includegraphics[width=1.0\linewidth]{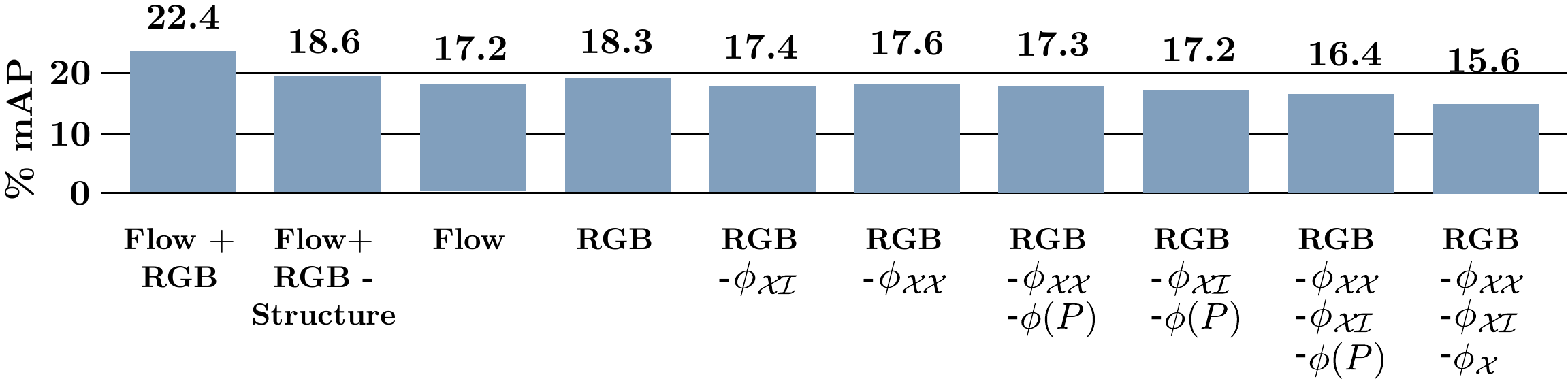}
\caption{Ablation analysis for our proposed model. Y-axis is video classification mAP \%. Each factor helps in improving the overall model performance. $\phi(P)$ indicates dropping the `progress' term within the semantic factor $\phi_{\frame}$ .}
\label{tbl:ablation}
\end{figure}

\myparagraph{Ablation studies} To study the contribution of different model parts, we also train ablated versions of our model separately choosing the best hyperparameters for each version. In addition to our model with only RGB or Flow, we also consider dropping $\phi_{\pair}$ (i.e., no sequential information), $\phi_{\intent}$ (i.e., no intent), both (i.e., only semantic information), and further dropping $\phi_{\frame}$ (i.e., dropping all structure). Figure~\ref{tbl:ablation} shows that semantic reasoning improves over the baseline. Further, while both $\phi_{\intent}$ and $\phi_{\pair}$ capture temporal information, they are complementary.

\begin{figure*}[th]
\centering
\includegraphics[width=1\linewidth]{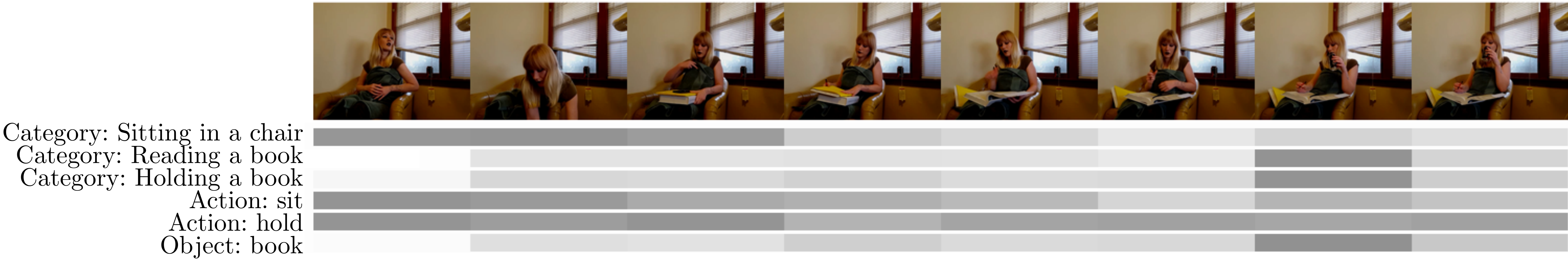}
\caption{Model predictions for a sample video. We see the interplay between categories, objects and actions over time. For example, model becomes confident about the action \emph{sit} early, which aids the understanding of \emph{Sitting in a chair} once the chair becomes visible, and helps predicting {\em Reading a book}. Darker colors represent higher likelihood, and we average predictions to correspond to each frame.}
\label{fig:examples}
\end{figure*}

\subsection{Temporal Localization}

To measure the ability of the methods to temporally localize and understand when exactly activities happen, we adapt the benchmark of~\cite{charades} to evaluate with the same mAP metric but on individual frames. That is, instead of having a single prediction per video, evaluation is now split into 25 equidistant timepoints having zero or more activities, and the models make a prediction for each of those\footnote{This evaluation code has been included as a part of the Charades dataset (\myurl{allenai.org/plato/charades/}).}.
We find this way of evaluating localization robust to annotation ambiguity, and informative for challenging datasets.
All hyperparameters were kept equal between localization and classification experiments. All baselines are run on 75 frames across the video, and then every third frame selected for a total of 25 frames. We also considered methods with \emph{post-processing} where the model predictions for the 75 frames are averaged across 30 frames to obtain more spatial consistency, and then 25 frames selected as before.
\begin{table}[tb]
\begin{center}
\setlength{\tabcolsep}{8pt}
\resizebox{\linewidth}{!}{%
\begin{tabular}{rcccccc} \toprule
 & \multirow{ 2}{*}{Random} & \multirow{ 2}{*}{RGB} & \multirow{ 2}{*}{Two-Stream++} & Two-Stream & Two-Stream & \multirow{ 2}{*}{Ours}  \\ 
 & & & &  +LSTM & Extended & \\ \hline \rule{0pt}{3ex}
\emph{Standard} & 2.42 & 7.89 & 8.94 & 9.60 & 9.37 & {\bf 9.69} \\
\emph{Post-processing} & 2.42 & 9.05 & 10.9 & 10.4 & 11.6 & {\bf 12.8} \\
\bottomrule
\end{tabular}
}
\end{center}
\caption{Temporal localization results (mAP \%) on the Charades~\cite{charades} dataset. Our proposed method outperforms the LSTM model, and is also more tractable to train at a large-scale.}
\label{tbl:frameclassification}
\end{table}

Table~\ref{tbl:frameclassification} shows that our method outperforms the alternatives, including the LSTM model which has been shown to be a powerful temporal modeling tool, but challenging to train on top of a two-stream network due to correlations between consecutive samples. These results demonstrate the our method is tractable way of training end-to-end structured models to understand activities. Interestingly, our method still benefits from adding post-processing, significantly more than the LSTM baseline, likely since our method is reasoning on larger time-scales. This suggests that our model could further benefit from joint training with additional kernels in the temporal term.

\myparagraph{Qualitative visualization} 
A key advantage of our model is the structured understanding of videos in terms of multiple aspects, such as action sequences, objects, and even intentions. 
To visualize this, we display predictions over time in Figure~\ref{fig:examples} for the three most confident activity categories, two most confident actions, and the most confident object. More examples are presented in the Appendix.

\begin{figure}
\centering
\includegraphics[width=0.9\linewidth]{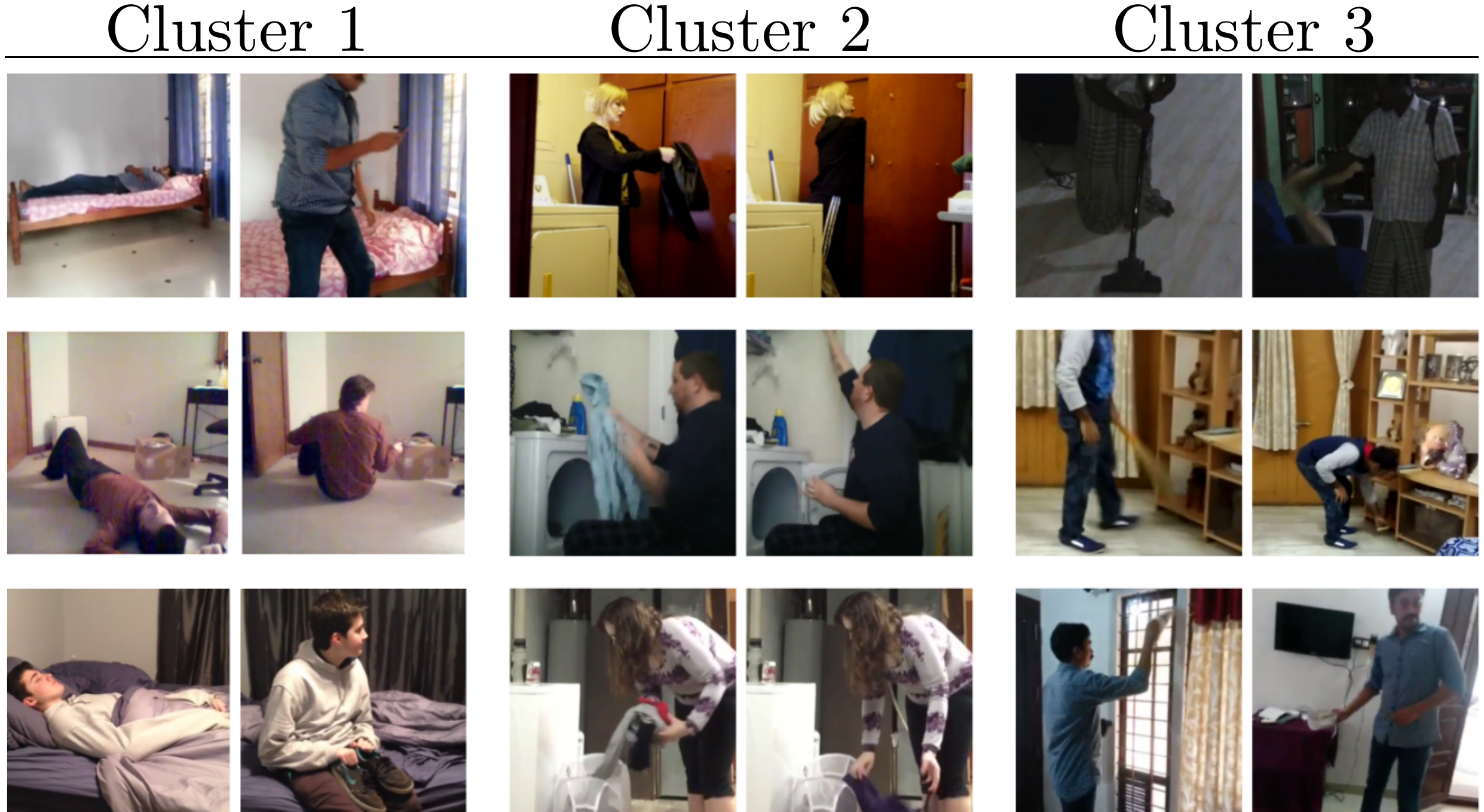}
\caption{To visualize the learned intent, we cluster videos based on intent. In \emph{Cluster 1}, the model captures the intent of get up from lying down. In \emph{Cluster 2}, folding clothes is followed by putting them away, and \emph{Cluster 3} shows cleaning with a broom/vacuum/towel, followed by picking up things.}
\label{fig:goals}
\end{figure}

\myparagraph{Interpretation of Intent}
In our model, the intent $I$ is a continuous distribution over the latent variables. To get an insight into how our model learns the intent, we ran a simple experiment that clustered videos in the dataset that have the most similar inferred intent distributions. The first cluster in Figure~\ref{fig:goals} shows the model captures the simple intent that the person intends to get up from lying down. In the videos, these actions are 10-20 seconds apart, demonstrating that the intent helps reason over large time scales.

\begin{figure}
\centering
\includegraphics[width=1.0\linewidth]{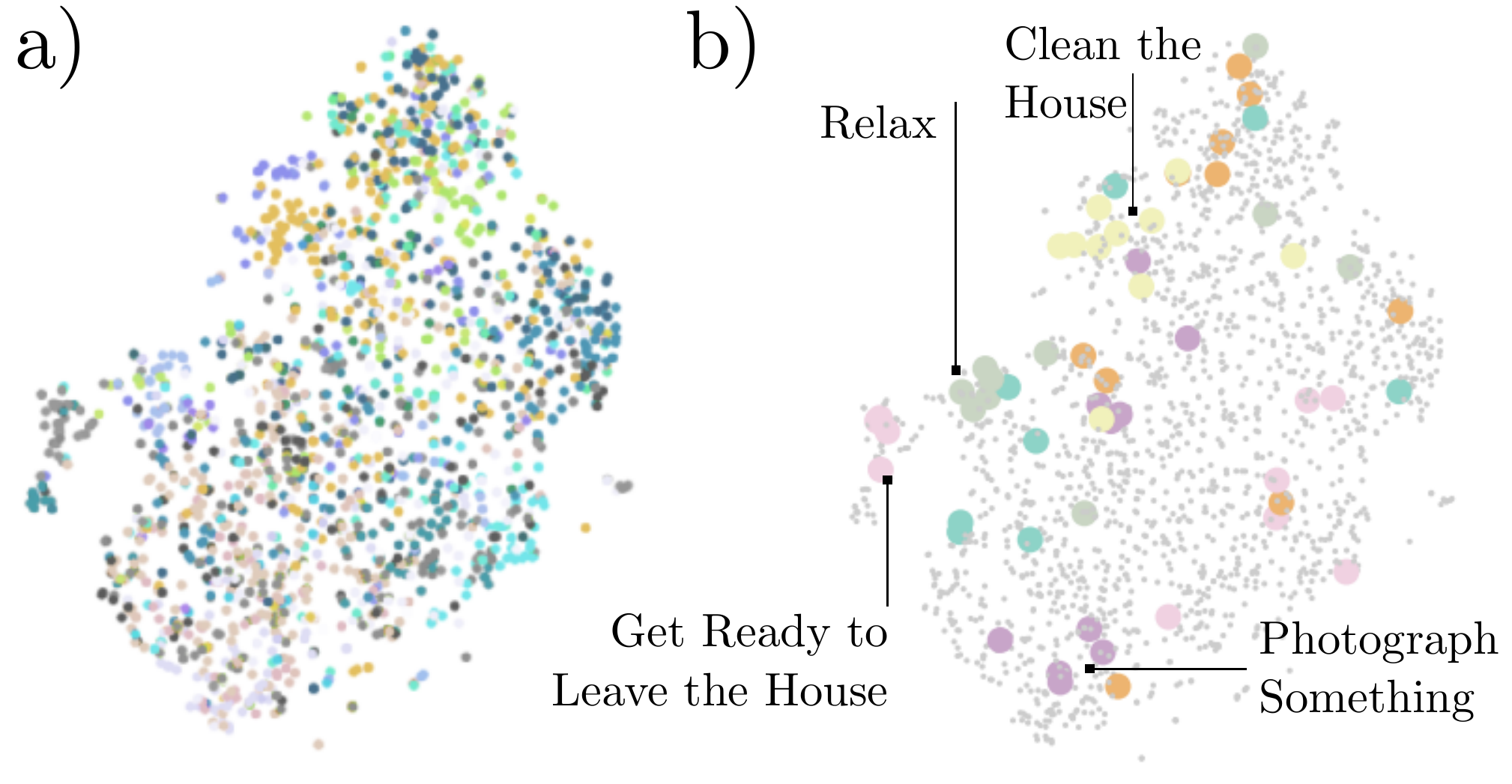}
\caption{t-SNE visualization for the learned intent. Each point corresponds to a video. In a) it is colored based on its activity shared by the most of the 10 nearest neighbors (each video has multiple actions). In b) videos with 6 annotated intent types are emphasized with larger points colored by the type.}
\label{fig:intentannotations}
\end{figure}

In order to further analyze the `intent' variable, we plot the t-SNE embedding of the intent variable for the videos in the test set. We see that there is clear clustering of similar videos in Fig.~\ref{fig:intentannotations}a. We also annotated 10 types of intent (100 videos total). More details are presented in the Appendix. We observe that the intent representation preserves some of the intent types in Fig.~\ref{fig:intentannotations}b. Quantitatively, even without mitigating outliers, the average distance (in $10^{-3}$) between pairs of videos within an intent type was $6.02$ compared to $7.25$ ($\sigma{=}1.06$) for any points, and the difference is significant for 5 of 10 intent types (p=$0.1$). This tentatively suggest that the intent captures interesting structure in the data, and we hope this will encourage future work.

\section{Conclusion} 

In this paper, we have presented a deep-structured model using a fully-connected temporal CRF that not only models semantic aspects of activities but also reasons about long-term temporal relations. We also presented an asynchronous stochastic inference algorithm that circumvents a key bottleneck in the large-scale end-to-end model learning. Using our proposed method, we have demonstrated impressive activity classification and temporal localization results on a challenging dataset of realistic activities. 

{\footnotesize
\noindent {\bf Acknowledgements:} This work was partly supported by ONR MURI N00014-16-1-2007, ONR N00014-13-1-0720, NSF IIS-1338054, NSF-1652052, NRI-1637479, Intel via the Intel Science and Technology Center for Visual Cloud Systems, Allen Distinguished Investigator Award, gifts from Google, and the Allen Institute for Artificial Intelligence. The authors would like to thank Mark Yatskar for lending his expertise on deep CRFs, and Olga Russakovsky, Christoph Dann, and the anonymous reviewers for their invaluable suggestions and advice.}

\balance

{\small
\bibliographystyle{ieee}
\bibliography{cvpr2017gunnar}

\begin{thebibliography}{10}\itemsep=-1pt

\bibitem{ChenSchwingICML2015}
L.-C. Chen$^\ast$, A.~G. Schwing$^\ast$, A.~L. Yuille, and R.~Urtasun.
\newblock {Learning Deep Structured Models}.
\newblock In {\em Proc. ICML}, 2015.
\newblock $^\ast$ equal contribution.

\bibitem{MBH06}
N.~Dalal, B.~Triggs, and C.~Schmid.
\newblock Human detection using oriented histograms of flow and appearance.
\newblock In {\em ECCV}, 2006.

\bibitem{Souza16}
C.~R. de~Souza, A.~Gaidon, E.~Vig, and A.~M. López.
\newblock Sympathy for the details: Dense trajectories and hybrid
  classification architectures for action recognition.
\newblock In {\em ECCV}, 2016.

\bibitem{Donahue15}
J.~Donahue, L.~A. Hendricks, S.~Guadarrama, M.~Rohrbach, S.~Venugopalan,
  K.~Saenko, and T.~Darrell.
\newblock Long-term recurrent convolutional networks for visual recognition and
  description.
\newblock In {\em CVPR}, 2015.

\bibitem{Fathi13}
A.~Fathi and J.~M. Rehg.
\newblock Modeling actions through state changes.
\newblock In {\em ICCV}, 2013.

\bibitem{Basura15}
B.~Fernando, E.~Gavves, M.~J. Oramas, A.~Ghodrati, and T.~Tuytelaars.
\newblock Modeling video evolution for action recognition.
\newblock In {\em CVPR}, 2015.

\bibitem{Georgia15}
G.~Gkioxari and J.~Malik.
\newblock Finding action tubes.
\newblock In {\em CVPR}, 2015.

\bibitem{THUMOS15}
A.~Gorban, H.~Idrees, Y.-G. Jiang, A.~Roshan~Zamir, I.~Laptev, M.~Shah, and
  R.~Sukthankar.
\newblock {THUMOS} challenge: Action recognition with a large number of
  classes.
\newblock \url{http://www.thumos.info/}, 2015.

\bibitem{gupta2009observing}
A.~Gupta, A.~Kembhavi, and L.~S. Davis.
\newblock Observing human-object interactions: Using spatial and functional
  compatibility for recognition.
\newblock {\em TPAMI}, 2009.

\bibitem{VSRL_gupta15}
S.~Gupta and J.~Malik.
\newblock Visual semantic role labeling.
\newblock {\em CoRR}, /abs/1505.04474, 2015.

\bibitem{hinton2002training}
G.~E. Hinton.
\newblock Training products of experts by minimizing contrastive divergence.
\newblock {\em Neural computation}, 14(8):1771--1800, 2002.

\bibitem{Izadinia12}
H.~Izadinia and M.~Shah.
\newblock Recognizing complex events using large margin joint low-level event
  model.
\newblock {\em ECCV}, 2012.

\bibitem{ArpitJ13}
A.~Jain, A.~Gupta, M.~Rodriguez, and L.~S. Davis.
\newblock Representing videos using mid-level discriminative patches.
\newblock In {\em CVPR}, 2013.

\bibitem{3DCNN}
S.~Ji, W.~Xu, M.~Yang, and K.~Yu.
\newblock 3d convolutional neural networks for human action recognition.
\newblock {\em TPAMI}, 2013.

\bibitem{Karpathy14}
A.~Karpathy, G.~Toderici, S.~Shetty, T.~Leung, R.~Sukthankar, and L.~Fei-Fei.
\newblock Large-scale video classification with convolutional neural networks.
\newblock In {\em CVPR}, 2014.

\bibitem{kitani2012activity}
K.~M. Kitani, B.~D. Ziebart, J.~A. Bagnell, and M.~Hebert.
\newblock Activity forecasting.
\newblock In {\em ECCV}, 2012.

\bibitem{HOG3D}
A.~Klaser, M.~Marszalek, and C.~Schmid.
\newblock A spatio-temporal descriptor based on 3d-gradients.
\newblock In {\em BMVC}, 2008.

\bibitem{koller}
D.~Koller and N.~Friedman.
\newblock {\em Probabilistic graphical models: principles and techniques}.
\newblock 2009.

\bibitem{densecrfsegmentation}
P.~Kr{\"a}henb{\"u}hl and V.~Koltun.
\newblock Efficient inference in fully connected crfs with gaussian edge
  potentials.
\newblock In {\em NIPS}, 2011.

\bibitem{alexnet12}
A.~Krizhevsky, I.~Sutskever, and G.~E. Hinton.
\newblock Imagenet classification with deep convolutional neural networks.
\newblock In {\em NIPS}, 2012.

\bibitem{lan2015iccv}
T.~Lan, Y.~Zhu, A.~R. Zamir, and S.~Savarese.
\newblock Action recognition by hierarchical mid-level action elements.
\newblock In {\em ICCV}, 2015.

\bibitem{Lan15}
Z.~Lan, M.~Lin, X.~Li, A.~G. Hauptmann, and B.~Raj.
\newblock Beyond gaussian pyramid: Multi-skip feature stacking for action
  recognition.
\newblock In {\em CVPR}, 2015.

\bibitem{STIP05}
I.~Laptev.
\newblock On space-time interest points.
\newblock {\em IJCV}, 64, 2005.

\bibitem{HOF}
I.~Laptev, M.~Marszalek, C.~Schmid, and B.~Rozenfeld.
\newblock Learning realistic human actions from movies.
\newblock In {\em CVPR}, 2008.

\bibitem{Le11}
Q.~V. Le, W.~Y. Zou, S.~Y. Yeung, and A.~Y. Ng.
\newblock Learning hierarchical invariant spatio-temporal features for action
  recognition with independent subspace analysis.
\newblock In {\em CVPR}, 2011.

\bibitem{li2007and}
L.-J. Li and L.~Fei-Fei.
\newblock What, where and who? classifying events by scene and object
  recognition.
\newblock In {\em ICCV}, 2007.

\bibitem{Matikainen09}
P.~Matikainen, M.~Hebert, and R.~Sukthankar.
\newblock Trajectons: Action recognition through the motion analysis of tracked
  features.
\newblock In {\em ICCV Workshops}, 2009.

\bibitem{Snoek16}
P.~Mettes, J.~C. van Gemert, and C.~G. Snoek.
\newblock Spot on: Action localization from pointly-supervised proposals.
\newblock In {\em ECCV}, 2016.

\bibitem{mnih2013playing}
V.~Mnih, K.~Kavukcuoglu, D.~Silver, A.~Graves, I.~Antonoglou, D.~Wierstra, and
  M.~Riedmiller.
\newblock Playing atari with deep reinforcement learning.
\newblock {\em arXiv preprint arXiv:1312.5602}, 2013.

\bibitem{Peng14}
X.~Peng, C.~Zou, Y.~Qiao, and Q.~Peng.
\newblock Action recognition with stacked fisher vectors.
\newblock In {\em ECCV}, 2014.

\bibitem{pirsiavash2014parsing}
H.~Pirsiavash and D.~Ramanan.
\newblock Parsing videos of actions with segmental grammars.
\newblock In {\em CVPR}, 2014.

\bibitem{pirsiavash2014inferring}
H.~Pirsiavash, C.~Vondrick, and A.~Torralba.
\newblock Inferring the why in images.
\newblock {\em arXiv preprint arXiv:1406.5472}, 2014.

\bibitem{poppe2010survey}
R.~Poppe.
\newblock A survey on vision-based human action recognition.
\newblock {\em Image and vision computing}, 28(6):976--990, 2010.

\bibitem{premack1978does}
D.~Premack and G.~Woodruff.
\newblock Does the chimpanzee have a theory of mind?
\newblock {\em Behavioral and brain sciences}, 1(04):515--526, 1978.

\bibitem{prest2012weakly}
A.~Prest, C.~Schmid, and V.~Ferrari.
\newblock Weakly supervised learning of interactions between humans and
  objects.
\newblock {\em TPAMI}, 2012.

\bibitem{Rohrbach12}
M.~Rohrbach, M.~Regneri, M.~Andriluka, S.~Amin, M.~Pinkal, and B.~Schiele.
\newblock Script data for attribute-based recognition of composite activities.
\newblock {\em ECCV}, 2012.

\bibitem{ryoo2007hierarchical}
M.~S. Ryoo and J.~Aggarwal.
\newblock Hierarchical recognition of human activities interacting with
  objects.
\newblock In {\em CVPR}, 2007.

\bibitem{Corso12}
S.~Sadanand and J.~J. Corso.
\newblock Action bank: A high-level representation of activity in video.
\newblock In {\em CVPR}, 2012.

\bibitem{salakhutdinov2009deep}
R.~Salakhutdinov and G.~E. Hinton.
\newblock Deep boltzmann machines.
\newblock In {\em AISTATS}, 2009.

\bibitem{schwing2015fully}
A.~G. Schwing and R.~Urtasun.
\newblock Fully connected deep structured networks.
\newblock {\em arXiv preprint arXiv:1503.02351}, 2015.

\bibitem{scnn_shou_wang_chang_cvpr16}
Z.~Shou, D.~Wang, and S.-F. Chang.
\newblock Temporal action localization in untrimmed videos via multi-stage
  cnns.
\newblock In {\em CVPR}, 2016.

\bibitem{sigurdsson2016learning}
G.~A. Sigurdsson, X.~Chen, and A.~Gupta.
\newblock Learning visual storylines with skipping recurrent neural networks.
\newblock In {\em ECCV}, 2016.

\bibitem{charades}
G.~A. Sigurdsson, G.~Varol, X.~Wang, A.~Farhadi, I.~Laptev, and A.~Gupta.
\newblock Hollywood in homes: Crowdsourcing data collection for activity
  understanding.
\newblock In {\em ECCV}, 2016.

\bibitem{Simonyan13}
K.~Simonyan, A.~Vedaldi, and A.~Zisserman.
\newblock Deep inside convolutional networks: Visualising image classification
  models and saliency maps.
\newblock {\em CoRR}, /abs/1312.6034, 2013.

\bibitem{2stream14}
K.~Simonyan and A.~Zisserman.
\newblock Two-stream convolutional networks for action recognition in videos.
\newblock In {\em NIPS}, 2014.

\bibitem{Simonyan15}
K.~Simonyan and A.~Zisserman.
\newblock Very deep convolutional networks for large-scale image recognition.
\newblock {\em ICLR}, 2015.

\bibitem{YaleS13}
Y.~Song, L.-P. Morency, and R.~Davis.
\newblock Action recognition by hierarchical sequence summarization.
\newblock In {\em CVPR}, 2013.

\bibitem{UCF101}
K.~Soomro, A.~Roshan~Zamir, and M.~Shah.
\newblock {UCF101}: A dataset of 101 human actions classes from videos in the
  wild.
\newblock In {\em CRCV-TR-12-01}, 2012.

\bibitem{Srivastava15}
N.~Srivastava, E.~Mansimov, and R.~Salakhutdinov.
\newblock Unsupervised learning of video representations using lstms.
\newblock {\em CoRR}, /abs/1502.04681, 2015.

\bibitem{Sun13}
C.~Sun and R.~Nevatia.
\newblock Active: Activity concept transitions in video event classification.
\newblock {\em ICCV}, 2013.

\bibitem{Sun_2015_ICCV}
L.~Sun, K.~Jia, D.-Y. Yeung, and B.~E. Shi.
\newblock Human action recognition using factorized spatio-temporal
  convolutional networks.
\newblock In {\em ICCV}, 2015.

\bibitem{KevinT12}
K.~Tang, L.~Fei-Fei, and D.~Koller.
\newblock Learning latent temporal structure for complex event detection.
\newblock In {\em CVPR}, 2012.

\bibitem{Taylor10}
G.~W. Taylor, R.~Fergus, Y.~LeCun, and C.~Bregler.
\newblock Convolutional learning of spatio-temporal features.
\newblock In {\em ECCV}, 2010.

\bibitem{Tran15}
D.~Tran, L.~Bourdev, R.~Fergus, L.~Torresani, and M.~Paluri.
\newblock Learning spatiotemporal features with 3d convolutional networks.
\newblock In {\em ICCV}, 2015.

\bibitem{Vondrick16_actns}
C.~Vondrick, D.~Oktay, H.~Pirsiavash, and A.~Torralba.
\newblock Predicting motivations of actions by leveraging text.
\newblock In {\em CVPR}, 2016.

\bibitem{Vondrick16_repr}
C.~Vondrick, H.~Pirsiavash, and A.~Torralba.
\newblock Anticipating visual representations from unlabeled video.
\newblock In {\em CVPR}, 2016.

\bibitem{WangIDT13}
H.~Wang and C.~Schmid.
\newblock Action recognition with improved trajectories.
\newblock In {\em ICCV}, 2013.

\bibitem{TDD15}
L.~Wang, Y.~Qiao, and X.~Tang.
\newblock Action recognition with trajectory-pooled deep-convolutional
  descriptors.
\newblock In {\em CVPR}, 2015.

\bibitem{Wang_Transformation}
X.~Wang, A.~Farhadi, and A.~Gupta.
\newblock Actions {\textasciitilde} transformations.
\newblock In {\em CVPR}, 2016.

\bibitem{weinland2011survey}
D.~Weinland, R.~Ronfard, and E.~Boyer.
\newblock A survey of vision-based methods for action representation,
  segmentation and recognition.
\newblock {\em Computer vision and image understanding}, 115(2):224--241, 2011.

\bibitem{weinzaepfel2016towards}
P.~Weinzaepfel, X.~Martin, and C.~Schmid.
\newblock Towards weakly-supervised action localization.
\newblock {\em arXiv preprint arXiv:1605.05197}, 2016.

\bibitem{Xu15}
Z.~Xu, Y.~Yang, and A.~G. Hauptmann.
\newblock A discriminative cnn video representation for event detection.
\newblock {\em CVPR}, 2015.

\bibitem{yatskarsituation}
M.~Yatskar, L.~Zettlemoyer, and A.~Farhadi.
\newblock Situation recognition: Visual semantic role labeling for image
  understanding.
\newblock {\em CVPR}, 2016.

\bibitem{yeung2015every}
S.~Yeung, O.~Russakovsky, N.~Jin, M.~Andriluka, G.~Mori, and L.~Fei-Fei.
\newblock Every moment counts: Dense detailed labeling of actions in complex
  videos.
\newblock {\em arXiv preprint arXiv:1507.05738}, 2015.

\bibitem{yeung2015end}
S.~Yeung, O.~Russakovsky, G.~Mori, and L.~Fei-Fei.
\newblock End-to-end learning of action detection from frame glimpses in
  videos.
\newblock {\em arXiv preprint arXiv:1511.06984}, 2015.

\bibitem{cnnlstm}
J.~Yue-Hei~Ng, M.~Hausknecht, S.~Vijayanarasimhan, O.~Vinyals, R.~Monga, and
  G.~Toderici.
\newblock Beyond short snippets: Deep networks for video classification.
\newblock In {\em CVPR}, 2015.

\bibitem{raqueldense2016}
Z.~Zhang, S.~Fidler, and R.~Urtasun.
\newblock Instance-level segmentation with deep densely connected mrfs.
\newblock In {\em CVPR}, 2016.

\end{thebibliography}
}


\newpage
\onecolumn
\section{Appendix}
This appendix contains the following additional content:

\begin{enumerate}
\item Description of the CRF.
\item Derivation of the update equations.
\item Details of the learning algorithm.
\item Additional implementation details.
\item Details about intent analysis.
\item Additional visualizations of output predictions.
\end{enumerate}

\subsection{Description of the CRF}

We create a CRF which predicts activity, object, etc., for every frame in the video. For reasoning about time, we create a \emph{fully-connected temporal CRF}, referred to as Asynchronous Temporal Field in the text. That is, unlike a linear-chain CRF for temporal modelling (the discriminative counterpart to Hidden Markov Models), each node depends on the state of every other node in the graph. We incorporate intention as another latent variable which is connected to all the action nodes.

In this work we encode multiple components of an activity. Each video with $T$ frames is represented as $\{X_1, \dots, X_T, I\}$ where $X_t$ is a set of frame-level random variables for time step $t$ and $I$ is a random variable that represent global intent in the entire video. As discussed in the paper, for clarity of derivation $X_t$ includes all frame level variables (${C_t,O_t,A_t,P_t,S_t}$)

Mathematically we consider a random field $\{X,I\}$ over all the random variables in our model ($\{X_1, \dots, X_T, I\}$). We now list the complete description of the CRF.

\paragraph{CRF Variables:}
\begin{itemize} 
\item Random field $\{X,I\} = \{X_1, \dots, X_T, I\}$
\item Frame $X_t = \{C_t,O_t,A_t,P_t,S_t\}$, $X_t \in \mathcal{X}, \mathcal{X} = \mathcal{C} {\times} \mathcal{O} {\times} \mathcal{A} {\times} \mathcal{P} {\times} \mathcal{S}$
\begin{itemize}
    \item Category $C_t \in \mathcal{C}, \mathcal{C}=\{1,2,...,157\}$ (For each category in the dataset)
    \item Object $O_t \in \mathcal{O}, \mathcal{O}=\{1,2,...,38\}$ (Includes "No object")
    \item Action $A_t \in \mathcal{A}, \mathcal{A}=\{1,2,...,33\}$
    \item Progress $P_t \in \mathcal{P}, \mathcal{P}=\{1,2,3\}$ (Before, Middle, End)
    \item Scene $S_t \in \mathcal{S}, \mathcal{S}=\{1,2,...,15\}$
\end{itemize}
\item Intent $I \in \mathcal{I}, \mathcal{I}=\{1,2,...,N_I\}$ ($N_I=30$ in this work)
\end{itemize}

\paragraph{CRF Potentials:}
\begin{itemize} 
\item $\phi_\frame \colon \mathcal{X} \mapsto \mathcal{R}$, equivalently: $\phi_\frame \colon \mathcal{C} {\times} \mathcal{O} {\times}  \mathcal{A} {\times} \mathcal{P} {\times}  \mathcal{S} \mapsto \mathcal{R}$
\item $\phi_\frame$ decomposes as follows: $\phi_\frame(C_t,O_t,A_t,P_t,S_t){=}\phi(O_t,P_t){+}\phi(A_t,P_t){+}\phi(O_t,S_t){+}\phi(C_t,O_t,A_t,P_t)$
\begin{itemize}
    \item $\phi(O_t,P_t) \colon \mathcal{O} {\times} \mathcal{P} \mapsto \mathcal{R}$
    \item $\phi(A_t,P_t) \colon \mathcal{A} {\times} \mathcal{P} \mapsto \mathcal{R}$
    \item $\phi(O_t,S_t) \colon \mathcal{O} {\times} \mathcal{S} \mapsto \mathcal{R}$
    \item $\phi(C_t,O_t,A_t,P_t) \colon \mathcal{B} \mapsto \mathcal{R}$, here $\mathcal{B}$ is all configurations of $C_t,O_t,A_t,P_t$ that exist in the training data.
\end{itemize}
\item $\phi_\intent \colon \mathcal{X}{\times} \mathcal{I} \mapsto \mathcal{R}$ (specifically we parametrize this as $\phi_\intent \colon \mathcal{O}{\times} \mathcal{I} \mapsto \mathcal{R}$)
\item $\phi_\pair \colon \mathcal{X}{\times} \mathcal{X} \mapsto \mathcal{R}$ (specifically we parametrize this as $\phi_\intent \colon \mathcal{O}{\times} \mathcal{O} \mapsto \mathcal{R}$)
\end{itemize}

\begin{figure*}[t]
\centering
\includegraphics[width=0.9\linewidth]{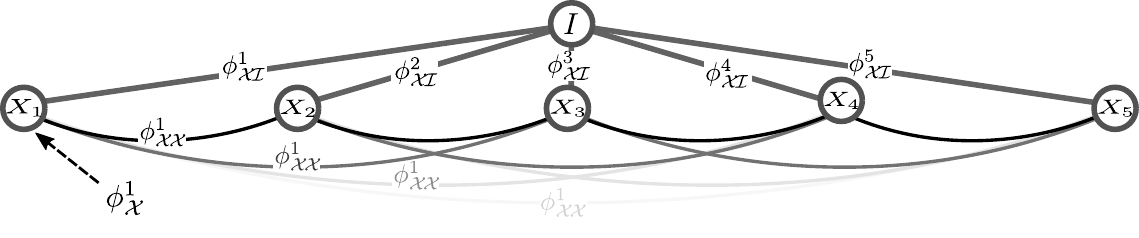}
\caption{The model captures interactions between all frames $X_t$ and the intent $I$, that is, a fully-connected model. Here shown for $T=5$. We visualize some of the potentials of the model, and where they fit into the graph. All $\phi^i_\intent$ share the same parameters, but we calculate the gradients with respect for each of them separately below. For efficient inference, we use a mean-field approximation presented below. A mean-field approximation is a simpler distribution that is fit to the original distribution when needed.}
\label{fig:modeldetail}
\end{figure*}

The complete distribution of the model is:
\begin{align}
P(X,I) = \frac{1}{Z} \exp \left\{ \sum_i \phi^i_\frame(x_i) + \sum_{i} \phi^i_\intent(x_i,I) + \sum_i \sum_{j \neq i} \phi^i_\pair(x_i,x_j) \right\}
\label{eq:complete}
\end{align}
where $\phi_\pair(x_i,x_j)$ is the potential between frame $i$ and frame $j$, and $\phi_\intent(x_i,I)$ is the potential between frame $i$ and the intent. For notational clarity $\phi_\frame(x_i)$ incorporates all potentials for ${C_t,O_t,A_t,P_t,S_t}$. The model is presented in Figure~\ref{fig:modeldetail}.

\subsection{Derivation of the Update Equations}
Given an input video $V{=}\{V_1,\dots,V_T\}$, our goal is to estimate the maximum a posteriori labeling of the random field by marginalizing over the intent $I$, $\sum_I P(X,I|V)$ as discussed in the paper. In the following derivations we omit the conditioning on $V$ and write $P(X,I)$ and $\phi(X,I)$.

Before we present the update equations and gradients, we define the following messages which will be used in the final version of the following equations for clarity in their presentation. Messages are a term used for cached computations sent between different functions in a dynamic programming fashion. In the following derivations, $X^*$ is used to explicitly denote the ground truth used for training. Plain $X$ is used to refer to the variable.

\noindent {\bf Outgoing Messages} (Messages that are calculated from a single frame)
\begin{align}
    F\!A_j(x_j) &= E_{U \sim Q_j} \left[ \mu(x_j,U) \right] \label{eq:firstoutmessage}\\
    F\!B_j(x_j) &= E_{U \sim Q_j} \left[ \mu(U,x_j) \right] \\
    H_j(I) &= E_{U \sim Q_j} \left[ \phi_\intent (U, I) \right] \\
    H^*_j(I) &= \phi_\intent(x_j^*,I) \\
    K_j(x_j) &= Q_j(x_j) \\
    K^*_j(x_j) &= \mathbf{1}_{x_j=x^*_j} \label{eq:lastoutmessage}
\end{align}

\noindent {\bf Incoming Messages} (Messages that are calculated from messages from multiple frames and used for the computation of a single frame)
\begin{align}
    \mathbb{F}\!\mathbb{A}_i(x_i) &= \sum_{j>i} E_{U_j \sim Q_j}[\mu(x_i,U_j)] K(v_i,v_j) = \sum_{j>i} F\!A_j(x_i)K(v_i,v_j) \label{eq:firstinmessage}\\
    \mathbb{F}\!\mathbb{B}_i(x_i) &= \sum_{j<i} E_{U_j \sim Q_j}[\mu(U_j,x_i)]K(v_j,v_i) = \sum_{j<i} F\!B_j(x_i)K(v_j,v_i) \\
    \mathbb{H}_i(I) &= \sum_{j \neq i} E_{U_j\sim Q_j} \left[ \phi_\intent (U_j,I) \right] = \sum_{j \neq i} H_j(I) \\
    \mathbb{H}^*_i(I) &= \sum_{j \neq i} \phi_\intent (x^*_j, I) = \sum_{j \neq i} H^*_j(I) \\
    \mathbb{K}\!\mathbb{A}_i(x_i) &= \sum_{j>i} Q_j(x_j) K(x_i,x_j) = \sum_{j>i} K_j(x_i) \\
    \mathbb{K}\!\mathbb{A}^*_i(x_i) &= \sum_{j>i} \mathbf{1}_{x_j=x^*_j}K(x_i,x^*_j) = \sum_{j>i} K^*_j(x_i) \\
    \mathbb{K}\!\mathbb{B}_i(x_i) &= \sum_{j<i} Q_j(x_j) K(x_j,x_i) = \sum_{j<i} K_j(x_i) \\
    \mathbb{K}\!\mathbb{B}^*_i(x_i) &= \sum_{j<i} \mathbf{1}_{x_j=x^*_j}K(x^*_j,x_i) = \sum_{j<i} K^*_j(x_i) \label{eq:lastinmessage}
\end{align}

Instead of computing the exact distribution $P(X,I)$ presented above, the structured variational approximation finds the distribution $Q(X,I)$ among a given family of distributions that best fits the exact distribution in terms of KL-divergence. By choosing a family of tractable distributions, it is possible to make inference involving the ideal distribution tractable. Here we use $Q(X,I)=Q_\qintent(I) \prod_i Q_i(x_i)$, the structured mean-field approximation. More details on mean-field approximation are presented section 11.5 generic update equation for $Q$ (Equation 11.54 in~\cite{koller}) is:
\begin{equation}
Q(x_i) \propto \exp \left\{ E_{X_{-i} \sim Q} \left[ \log P(x_i | X_{-i}) \right]\right\}
\label{eq:update}
\end{equation}
where $X_{-i}$ refers to all variables except $x_i$. Using Eq.~\ref{eq:complete} along with Eq.~\ref{eq:update} we get the following update equations:
\begin{align}
Q_i(x_i) \propto \exp \bigg\{ & \phi_\frame(x_i) + \mathrm{E}_{U\sim Q_\qintent} \left[ \phi_\intent(x_i,U) \right] \nonumber  
+ \sum_{j > i} \mathrm{E}_{U_j\sim Q_j} \left[ \phi_{\pair}(x_i,U_j) \right] \nonumber
+ \sum_{j < i} \mathrm{E}_{U_j\sim Q_j} \left[ \phi_{\pair}(U_j,x_i) \right] \bigg\} \nonumber \\
\propto \exp \bigg\{ & \phi_\frame(x_i) + \mathrm{E}_{U\sim Q_\qintent} \left[ \phi_\intent(x_i,U) \right] + \mathbb{F}\!\mathbb{A}_i(x_i) + \mathbb{F}\!\mathbb{B}_i(x_i) \bigg\} \\
Q_\qintent(I) \propto \exp \bigg\{ & \sum_j \mathrm{E}_{U_j\sim Q_j} \left[ \phi_\intent(U_j,I) \right] \bigg\} \\
\propto \exp \bigg\{ & \mathbb{H}_i(I) + H_i(I) \bigg\} \hspace{1cm} \text{(Here $i$ refers to the frame of interest, but any choice of $i$ holds)}
\label{eq:Q2}
\end{align}
where $Q_i$ is marginal distribution with respect to each of the frames, and $Q_\qintent$ is the marginal with respect to the intent.

\subsection{Details of the learning algorithm}

Training a deep CRF model requires calculating derivatives of the objective in terms of each of the potentials in the model, which in turn requires inference of $P(X,I|V)$. The network is trained to maximize the log-likelihood of the data:
\begin{align}
l(X^*) &= \log \sum_I P(X^*,I|V) \\
&= \log \sum_I \frac{\tilde{P}(X^*,I|V)}{Z(V)} \\
&= \log \sum_I \tilde{P}(X^*,I|V) - \log Z(V) \\
Z(V) &= \sum_I \sum_X \tilde{P}(X,I|V)
\end{align}
where we explicitly write out the partition function Z(V), and $\tilde{P}()$ is the unnormalized version of $P()$. Again, we use $X^*$ to explicitly refer to the ground truth labels. As before, $V$ is omitted from the following derivations.
The goal is to update the parameters of the model, for which we need gradients with respect to the parameters. Similar to SGD, we find the gradient with respect to one part of the parameters at a time, specifically with respect to one potential in one frame. That is, $\phi_\frame^i(x)$ instead of $\phi_\frame(x)$.
The partial derivatives of this loss with respect to each of the potentials are as follows.

\subsubsection{Updating the frame potential $\boldsymbol{\phi_\frame}$}

The frame potential $\phi_\frame(x_i)$ incorporates the interplay between activity category, object, action, progress and scene, and could be written explicitly as $\phi_\frame(C_t,O_t,A_t,P_t,S_t)$. In practice this potential is composed of unary, pairwise, and tertiary potentials directly predicted by a CNN. We found predicting only the following terms to be sufficient without introducing too many additional parameters:
$\phi_\frame(C_t,O_t,A_t,P_t,S_t){=}\phi(O_t,P_t){+}\phi(A_t,P_t){+}\phi(O_t,S_t)+ \phi(C_t,O_t,A_t,P_t)$ where we only model the assignments seen in the training set, and assume others are not possible.

Let us first derive the update equation for $\phi_\frame$ as a whole, and then demonstrate how to update each of the individual potentials. In the following derivation, we simply take the partial derivative where appropriate and iteratively use the chain rule.

\begin{align}
\frac{\partial l(X^*)}{\partial \phi_\frame^{\hat{i}} (\hat{x})} &= 
\frac{1}{\sum_I \tilde{P}(X^*,I)} \left( \sum_I \tilde{P}(X^*,I) \right) \frac{\partial \left( \sum_i \phi_\frame^i(x_i^*) \right) }{\partial \phi_\frame^{\hat{i}} (\hat{x}) }
 - \frac{\partial \log Z}{\partial \phi_\frame^{\hat{i}}(\hat{x})} \\
&= \mathbf{1}_{\hat{x}=x^*} - \frac{1}{Z} \sum_{X} \sum_I \frac{\partial \tilde{P}(X,I)}{\partial \phi_\frame^{\hat{i}}(\hat{x})} \hspace{1cm} \text{(Denominator and numerator cancel)}\\
&= \mathbf{1}_{\hat{x}=x^*} - \frac{1}{Z} \sum_{X} \sum_I \mathbf{1}_{\hat{x}=x} \tilde{P}(X,I) \\
&= \mathbf{1}_{\hat{x}=x^*} - \sum_{X} \sum_I \mathbf{1}_{\hat{x}=x} {P}(X,I) \\
&\approx \mathbf{1}_{\hat{x}=x^*} - \sum_{X} \sum_I \mathbf{1}_{\hat{x}=x} {Q}(X,I) \hspace{1cm} \text{(Using the mean-field)}\\
&= \mathbf{1}_{\hat{x}=x^*} - \sum_{X} \sum_I \mathbf{1}_{\hat{x}=x} Q_\qintent(I) \prod_i Q_i(x_i) \\
&= \mathbf{1}_{\hat{x}=x^*} - Q_{\hat{i}}(\hat{x}) \hspace{1cm} \text{(Since $\sum_{x_i} Q_i(x_i) = 1$)} \label{eq:finalgradients1}
\end{align}
where we use $X^*$ to refer to the ground truth labels, and $\hat{X}$ to refer to the variables we are taking the partial derivative with respect to. We note that $\frac{\partial \left( \sum_i \phi_\frame^i(x_i^*) \right) }{\partial \phi_\frame^{\hat{i}} (\hat{x})} = \mathbf{1}_{\hat{x}=x^*}$. Intuitively this implies the partial gradient is the difference between the ground truth and the model prediction.
This equation is easily extended to update each of the individual potentials as follows:
\begin{align}
\frac{\partial l(X^*)}{\partial \phi^{\hat{i}}(\hat{O}_t,\hat{P}_t)} &= 
\mathbf{1}_{(\hat{O}_t,\hat{P}_t)=(O^*_t,P^*_t)} - \sum_{C_t} \sum_{A_t} \sum_{S_t} Q_{\hat{i}}(X^*_t)\\
\frac{\partial l(X^*)}{\partial \phi^{\hat{i}}(\hat{A}_t,\hat{P}_t)} &= 
\mathbf{1}_{(\hat{A}_t,\hat{P}_t)=(A^*_t,P^*_t)} - \sum_{C_t} \sum_{O_t} \sum_{S_t} Q_{\hat{i}}(X^*_t)\\
\frac{\partial l(X^*)}{\partial \phi^{\hat{i}}(\hat{O}_t,\hat{S}_t)} &= 
\mathbf{1}_{(\hat{O}_t,\hat{S}_t)=(O^*_t,S^*_t)} - \sum_{C_t} \sum_{A_t} \sum_{P_t} Q_{\hat{i}}(X^*_t)\\
\frac{\partial l(X^*)}{\partial \phi^{\hat{i}}(\hat{C}_t,\hat{O}_t,\hat{A}_t,\hat{P}_t)} &= 
\mathbf{1}_{(\hat{C}_t,\hat{O}_t,\hat{A}_t,\hat{P}_t)=(C^*_t,O^*_t,A^*_t,P^*_t)} - \sum_{S_t} Q_{\hat{i}}(X^*_t)
\end{align}
where we marginalize out the variables that are not a part of each potential. Again, $X_t$ incorporates all the frame variables $\{C_t,O_t,A_t,P_t,S_t\}$.
These partial derivatives are passed down the CNN (backprop) to update the parameters of the network.

\subsubsection{Updating the frame-intent potential $\boldsymbol{\phi_\intent}$}

Similarly to $\phi_\frame$ we proceed as follows:
\begin{align}
\frac{\partial l(X^*)}{\partial \phi_\intent^{\hat{i}}(\hat{x},\hat{I})} &= 
\frac{1}{\sum_I \tilde{P}(X^*,I)} \left( \sum_I \tilde{P}(X^*,I) 
\mathbf{1}_{\hat{x}=x^*} \mathbf{1}_{\hat{I}=I} \right) - \frac{\partial \log Z}{\partial \phi_\intent^{\hat{i}}(\hat{x},\hat{I})} \\
&= \frac{\tilde{P}(X^*,\hat{I})}{\sum_I \tilde{P}(X^*,I)} 
\mathbf{1}_{\hat{x}=x^*}  - \frac{\partial \log Z}{\partial \phi_\intent^{\hat{i}}(\hat{x},\hat{I})} \\
&= \frac{\exp \left\{ \sum_i \phi_\intent^i(x^*_i,\hat{I}) \right\}}{\sum_I \exp \left\{ \sum_i \phi_\intent^i(x^*_i,I) \right\}} 
\mathbf{1}_{\hat{x}=x^*}  - \frac{\partial \log Z}{\partial \phi_\intent^{\hat{i}}(\hat{x},\hat{I})} \hspace{1cm} \text{(Terms without $I$ cancel)}\\
&= \frac{\exp \left\{ \sum_i \phi_\intent^i(x^*_i,\hat{I}) \right\}}{\sum_I \exp \left\{ \sum_i \phi_\intent^i(x^*_i,I) \right\}} 
\mathbf{1}_{\hat{x}=x^*}  - \frac{1}{Z} \sum_{X} \sum_I \frac{\partial \tilde{P}(X,I)}{\partial \phi_\intent^{\hat{i}}(\hat{x},\hat{I})} \\
&= \frac{\exp \left\{ \sum_i \phi_\intent^i(x^*_i,\hat{I}) \right\}}{\sum_I \exp \left\{ \sum_i \phi_\intent^i(x^*_i,I) \right\}} 
\mathbf{1}_{\hat{x}=x^*}  - \frac{1}{Z} \sum_{X} \sum_I \tilde{P}(X,I) \mathbf{1}_{\hat{x}=x} \mathbf{1}_{\hat{I}=I}\\
&= \frac{\exp \left\{ \sum_i \phi_\intent^i(x^*_i,\hat{I}) \right\}}{\sum_I \exp \left\{ \sum_i \phi_\intent^i(x^*_i,I) \right\}} 
\mathbf{1}_{\hat{x}=x^*}  - \sum_{X} \sum_I {P}(X,I) \mathbf{1}_{\hat{x}=x} \mathbf{1}_{\hat{I}=I}\\
&\approx \frac{\exp \left\{ \sum_i \phi_\intent^i(x^*_i,\hat{I}) \right\}}{\sum_I \exp \left\{ \sum_i \phi_\intent^i(x^*_i,I) \right\}} 
\mathbf{1}_{\hat{x}=x^*}  - \sum_{X} \sum_I {Q}(X,I) \mathbf{1}_{\hat{x}=x} \mathbf{1}_{\hat{I}=I} \hspace{1cm} \text{(Mean-field approximation)}\\
&= \frac{\exp \sum_i \phi_\intent(x^*_i,\hat{I})}{\sum_I \exp \sum_i  \phi_\intent(x^*_i,I)}\mathbf{1}_{\hat{x}=x^*} - Q_{\hat{i}}(\hat{x}) Q_\qintent(\hat{I})\\
&= \frac{\exp \left\{ \mathbb{H}^*_{\hat{i}}(\hat{I}) + H^*_{\hat{i}}(\hat{I}) \right\} }{\sum_I \exp \left\{ \mathbb{H}^*_i(I) + H^*_i(I) \right\} }\mathbf{1}_{\hat{x}=x^*} - Q_{\hat{i}}(\hat{x}) Q_\qintent(\hat{I})
\label{eq:finalgradients2}
\end{align}

This equation can be interpreted in that it captures the difference between the distribution of the intent given the ground truth, and the predicted distribution of the intent.

\subsubsection{Updating the frame-frame potential $\boldsymbol{\phi_\pair}$}

The pairwise potentials $\phi_{\pair}(x_i,x_j)$ for two time points $i$ and $j$ in our model have the form:
\begin{align}
\phi_{\pair}(x_i,x_j) &= \mu(x_i,x_j)\sum_{m} w^{(m)}k^{(m)}(v_i,v_j) \\
&= \mu(x_i,x_j) k(v_i,v_j)
\end{align}
where $\mu$ models the asymmetric affinity between frames, $w$ are kernel weights, and each $k^{(m)}$ is a Gaussian kernel that depends on the videoframes $v_i$ and $v_j$ which are omitted from this notation for convenience, but the probability and the potentials are conditioned on V. In this work we use a single kernel that prioritises short-term interactions:
\begin{align}
\label{eq:sigma2}
k(v_i,v_j) = \exp \left( - \frac{(j-i)^2}{2\sigma^2}\right)
\end{align}
The parameters of the general asymmetric compatibility function $\mu(x_i,x_j)$ are learned from the data, and $\sigma$ is a hyper-parameter chosen by cross-validation.
The parameters of $\mu$ are learned as follows, and this could be extended to a more general form of $\phi_{\pair}$:
\begin{align}
\frac{\partial l(X^*)}{\partial \mu^{\hat{i}}(\hat{x},\hat{b})}
&= \frac{1}{\sum_I \tilde{P}(X^*,I)} \left( \sum_I \tilde{P}(X^*,I) \right) \frac{\partial}{\partial \mu^{\hat{i}}(\hat{x},\hat{b})} \left( \sum_{j>\hat{i}} \phi_\pair^i(x^*_i,x^*_j) + \sum_{j<\hat{i}} \phi_\pair^i(x^*_j,x^*_i) \right) - \frac{\partial \log Z}{\partial \mu^{\hat{i}}(\hat{x},\hat{b})} \\
&= \sum_{j>\hat{i}} \mathbf{1}_{\hat{x}=x^*}\mathbf{1}_{\hat{b}=x^*_j}k(v_{\hat{i}},v_j)
+ \sum_{j<\hat{i}} \mathbf{1}_{\hat{x}=x^*}\mathbf{1}_{\hat{b}=x^*_j}k(v_j,v_{\hat{i}})
- \frac{1}{Z} \sum_{X} \sum_I \frac{\partial \tilde{P}(X,I)}{\partial \mu^{\hat{i}}(\hat{x},\hat{b})} \\
&= \sum_{j>\hat{i}} \mathbf{1}_{\hat{x}=x^*}\mathbf{1}_{\hat{b}=x^*_j}k(v_{\hat{i}},v_j)
+ \sum_{j<\hat{i}} \mathbf{1}_{\hat{x}=x^*}\mathbf{1}_{\hat{b}=x^*_j}k(v_j,v_{\hat{i}}) \nonumber \\
&\hspace{1em} - \frac{1}{Z} \sum_{X} \sum_I \tilde{P}(X,I) \sum_i \left( \sum_{j>i} \mathbf{1}_{\hat{x}=x}\mathbf{1}_{\hat{b}=x_j} k(v_i,v_j) + \sum_{j<i} \mathbf{1}_{\hat{x}=x}\mathbf{1}_{\hat{b}=x_j} k(v_j,v_i) \right)\\
&= \sum_{j>\hat{i}} \mathbf{1}_{\hat{x}=x^*}\mathbf{1}_{\hat{b}=x^*_j}k(v_{\hat{i}},v_j)
+ \sum_{j<\hat{i}} \mathbf{1}_{\hat{x}=x^*}\mathbf{1}_{\hat{b}=x^*_j}k(v_j,v_{\hat{i}}) \nonumber \\
&\hspace{1em} - \sum_{X} \sum_I Q_\qintent(I) \prod_i Q_i(x_i) \sum_i \left( \sum_{j>i} \mathbf{1}_{\hat{x}=x}\mathbf{1}_{\hat{b}=x_j} k(v_i,v_j) + \sum_{j<i} \mathbf{1}_{\hat{x}=x}\mathbf{1}_{\hat{b}=x_j} k(v_j,v_i) \right) \hspace{1cm} \text{(Mean-field)}\\
\frac{\partial l(X^*)}{\partial \mu^{\hat{i}}(a,b)} 
&= \sum_{j>\hat{i}} \mathbf{1}_{a=x^*_{\hat{i}}} \mathbf{1}_{b=x^*_j} k(v_{\hat{i}},v_j) - Q_{\hat{i}}(a) \sum_{j>\hat{i}} Q_j(b) k(v_{\hat{i}},v_j) 
+ \sum_{j<\hat{i}} \mathbf{1}_{b=x^*_{\hat{i}}} \mathbf{1}_{a=x^*_j} k(v_j,v_{\hat{i}}) - Q_{\hat{i}}(b) \sum_{j<\hat{i}} Q_j(a) k(v_j,v_{\hat{i}}) \\
&= \mathbf{1}_{a=x^*_{\hat{i}}} \mathbb{K}\!\mathbb{A}^*_{\hat{i}}(b) - Q_{\hat{i}}(a) \mathbb{K}\!\mathbb{A}_{\hat{i}}(b)
+ \mathbf{1}_{b=x^*_{\hat{i}}} \mathbb{K}\!\mathbb{B}^*_{\hat{i}}(a) - Q_{\hat{i}}(b) \mathbb{K}\!\mathbb{B}_{\hat{i}}(a)
\label{eq:finalgradients3}
\end{align}

This update equation consists of two symmetric parts, one for influence from frames before, and one for influence from frames after. Intuitively, this captures the difference in the true affinity between frame $i$ and all frames $j$ on the one hand, and on the other hand the predicted affinity, where the affinity is weighted by the kernel.

\subsection{Additional implementation details}

A more detailed algorithmic description of the model is presented in Algorithm~\ref{alg:full}. More details can be found on the project page \url{https://github.com/gsig/temporal-fields/}.

\begin{algorithm}
  \caption{Learning for Asynchronous Temporal Fields (Detailed)}
  \begin{algorithmic}[1]
      \State Given videos $\mathcal{V}$ 
      \While{$\textrm{not converged}$}
        \ForEach{example in mini-batch}
          \State Sample frame $v \in \mathbf{V}\subseteq \mathcal{V}$ that has index $i$
          \State Calculate messages with Eq.~\ref{eq:firstinmessage}-\ref{eq:lastinmessage}, approximated by Eq.~9 (from paper)
          \State Alternate updating $Q_i$ and $Q_\qintent$ until convergence
          \State Find gradients with Eqs.~\ref{eq:finalgradients1},\ref{eq:finalgradients2},\ref{eq:finalgradients3}
          \State Backprop gradients through CNN
          \State Store computations of Eq.~\ref{eq:firstoutmessage}-\ref{eq:lastoutmessage} for later use
        \EndFor
        \State Update CNN using accumulated gradients
      \EndWhile
  \end{algorithmic}
  \label{alg:full}
\end{algorithm}

\noindent {\bf Training time} Training the models in this paper took a while: The RGB stream of the Two-Stream model converged after only $0.2$ epochs ($20\%$ of the total data, randomly selected) of the training data, but training the Flow stream needed $4.0$ epochs to reach the best performance. Our model needed $0.7$ epochs for the RGB stream and $8.3$ epochs for the Flow stream. Each $0.1$ epoch is approximately $1450$ batches of size $256$ (all labelled frames at 8 FPS), and takes between 3-8 hours depending on hardware and model. Our learning rate schedule was chosen by finding the largest learning rate that did not cause divergence, and then making sure the learning rate was decayed by a factor of $100$ over the course of training. Investigations into training these kinds of models faster are likely to yield substantial benefits.

\noindent {\bf Training Deep Models with Latent Variables} One of the pursuits of this work was introducing latent variables into a deep framework, the intent. The gradient for the frame-intent potential, contains predictions of the model on both sides, which is a common problem in deep reinforcement learning, where a variety of tricks such as target fixing, double Q-learning, and gradient clipping, are used to combat the instability caused by this. In this work we found that simply severing the dependency of the frame-intent variable on the input data got rid of the instability, and still gave acceptable performance on the RGB stream, however we found that this did not give good performance on the Flow stream.

In order to train the network with the frame-intent potential depending on the input data, we experimented with a variety of techniques from the reinforcement learning literature. Only two methods were found to help: Alternating target and prediction networks, and regularization. For alternating target and prediction networks, the network predicts two frame-intent potentials, and then the network randomly chooses which to use as the target, and which to use as the source, and backprop only through one of them. For regularization, we enforce the frame-intent potential to be close to zero, similar to weight decay (set to $4\cdot10^{-4}$). Regularization was found to be give slightly better performance, and easy to implement/tune, and was used in this work.

\subsection{Details about intent analysis}

To analyze the learned intent variable, we defined 10 types of intent: \emph{getting something to eat}, \emph{clean the living space}, \emph{getting dressed}, \emph{getting something from storage}, \emph{get informed}, \emph{get out of bed}, \emph{leave the house}, \emph{photograph something}, \emph{relaxing}, \emph{working}. To identify videos corresponding to the intent, we used keyword related to the intent (such as \texttt{closet} and \texttt{clothes} for \emph{getting dressed}) and manually verified that the content of the video matched the intent. The analysis demonstrates that the latent intent variables captures non-trivial structure of the label space, but precisely understanding goal-oriented behavior compared to simple activity analysis remains important future work.

\subsection{Additional Visualizations of Output Predictions}

\begin{figure}[p]
\centering
\includegraphics[trim=0 30 0 0,clip,width=1.0\linewidth]{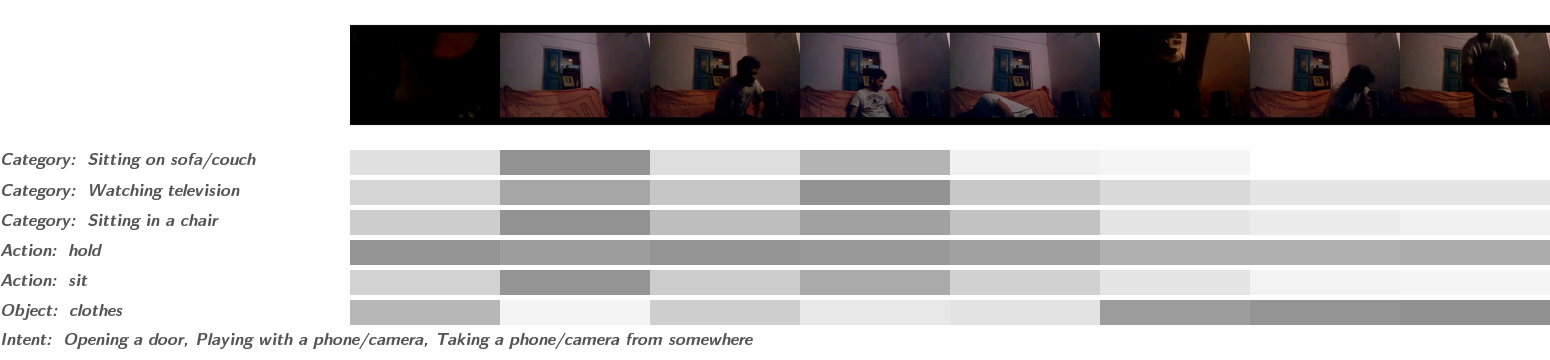}
\includegraphics[trim=0 30 0 0,clip,width=1.0\linewidth]{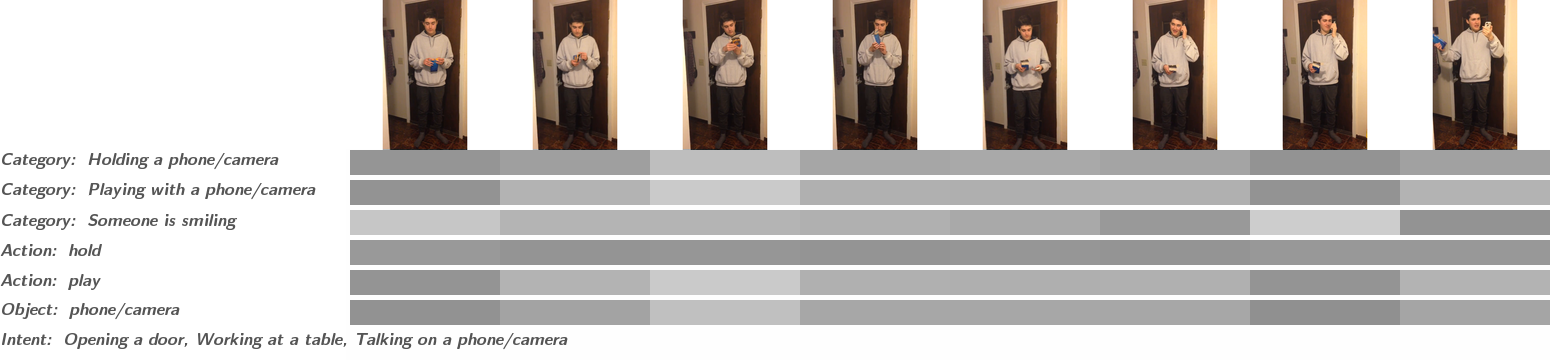}
\includegraphics[trim=0 30 0 0,clip,width=1.0\linewidth]{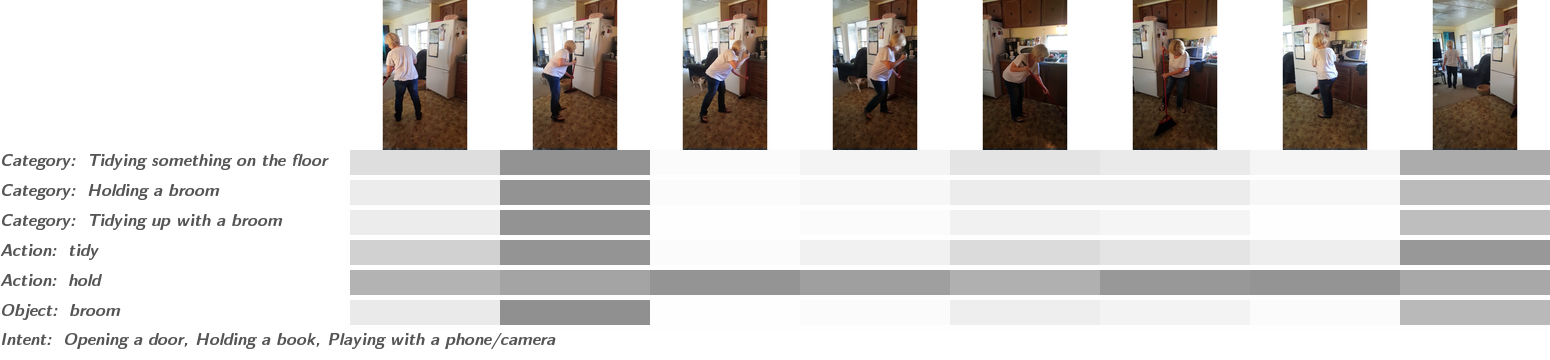}
\includegraphics[trim=0 30 0 0,clip,width=1.0\linewidth]{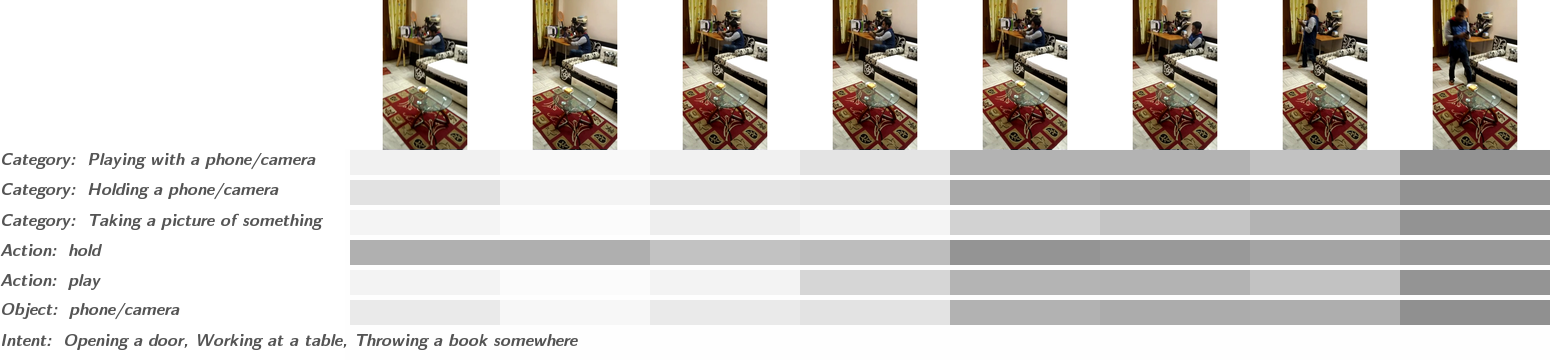}
\includegraphics[trim=0 30 0 0,clip,width=1.0\linewidth]{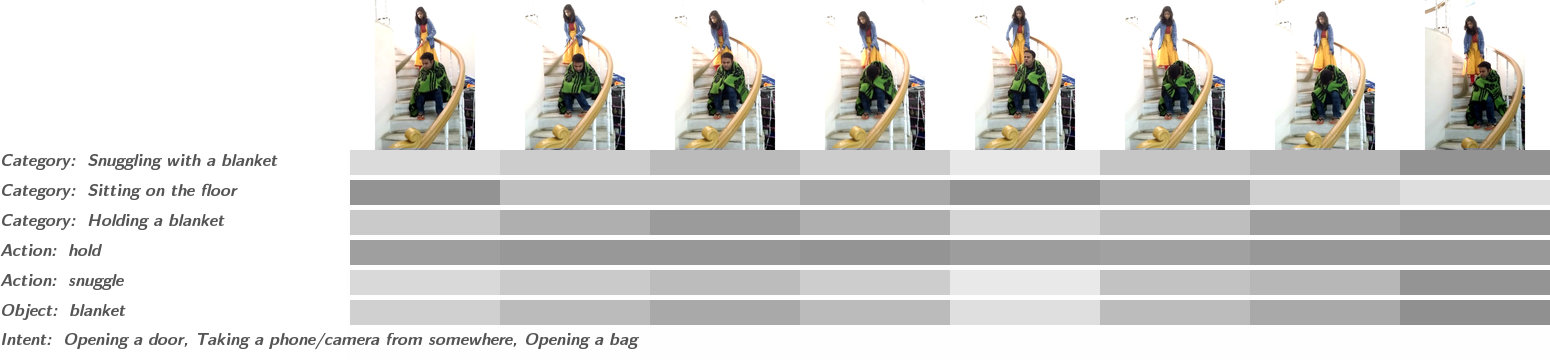}
\caption{Visualizations of the model predictions for the 3 most confident categories, 2 most confident actions, and 1 most confident object. Darker colors indicate higher likelihood.}
\label{fig:shortlist}
\end{figure}

\begin{figure}[p]
\centering
\includegraphics[trim=0 30 0 0,clip,width=1.0\linewidth]{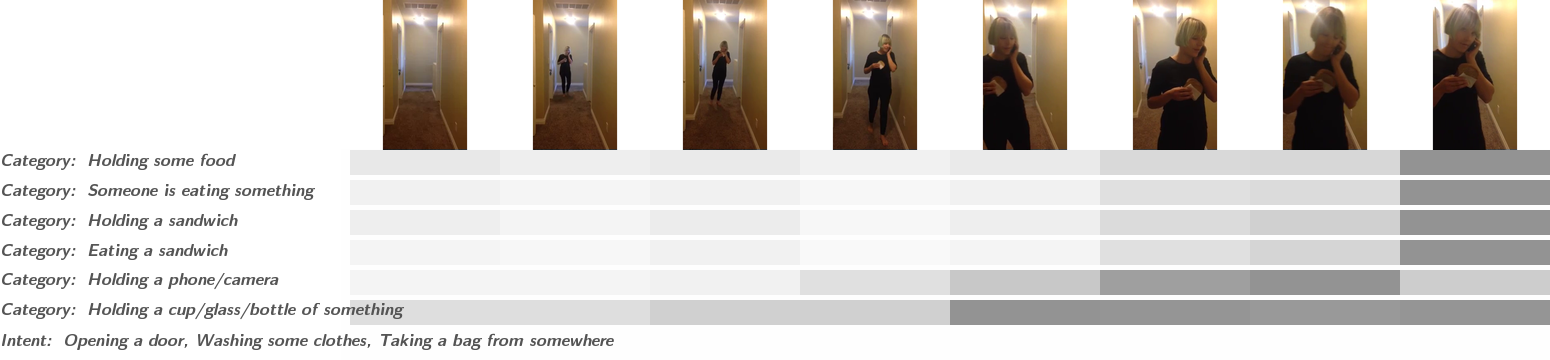}
\includegraphics[trim=0 30 0 0,clip,width=1.0\linewidth]{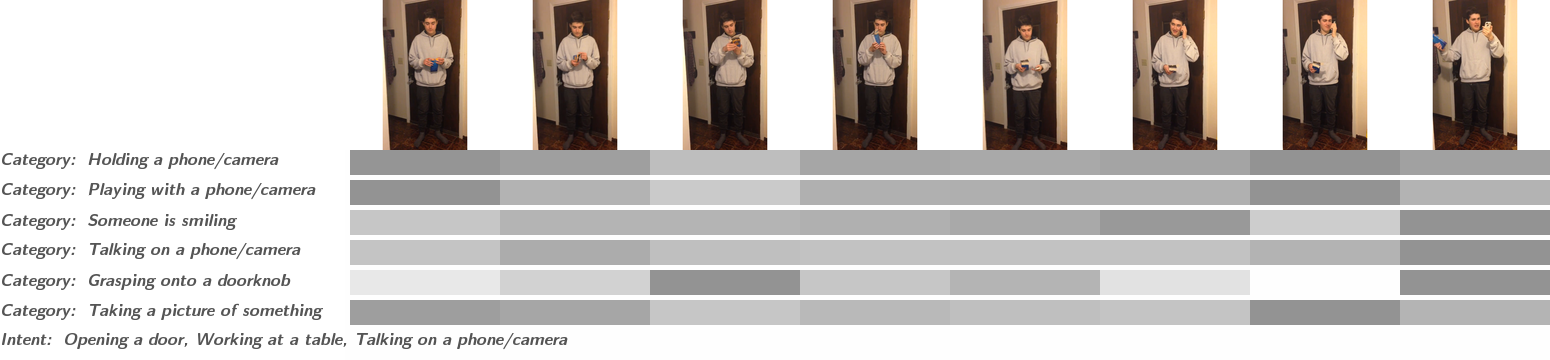}
\includegraphics[trim=0 30 0 0,clip,width=1.0\linewidth]{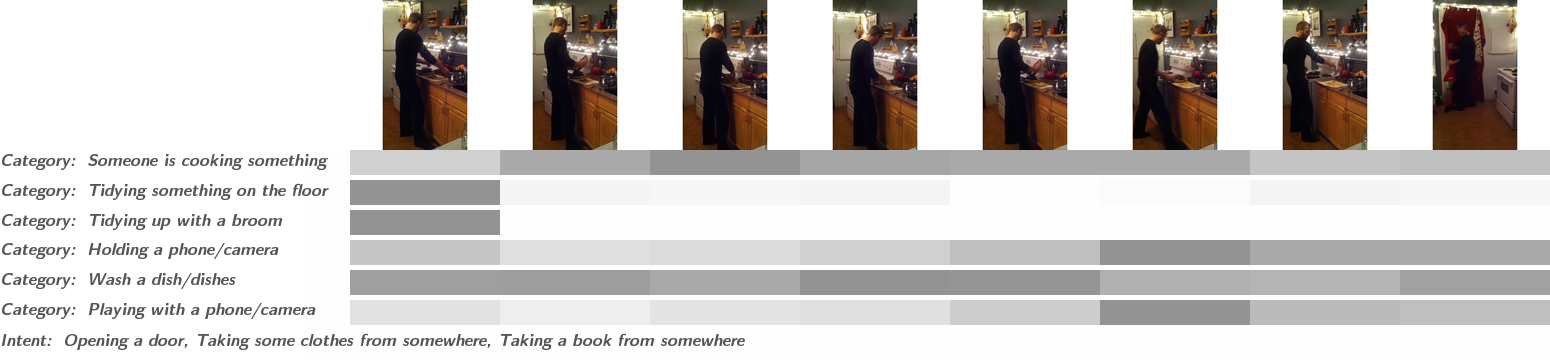}
\includegraphics[trim=0 30 0 0,clip,width=1.0\linewidth]{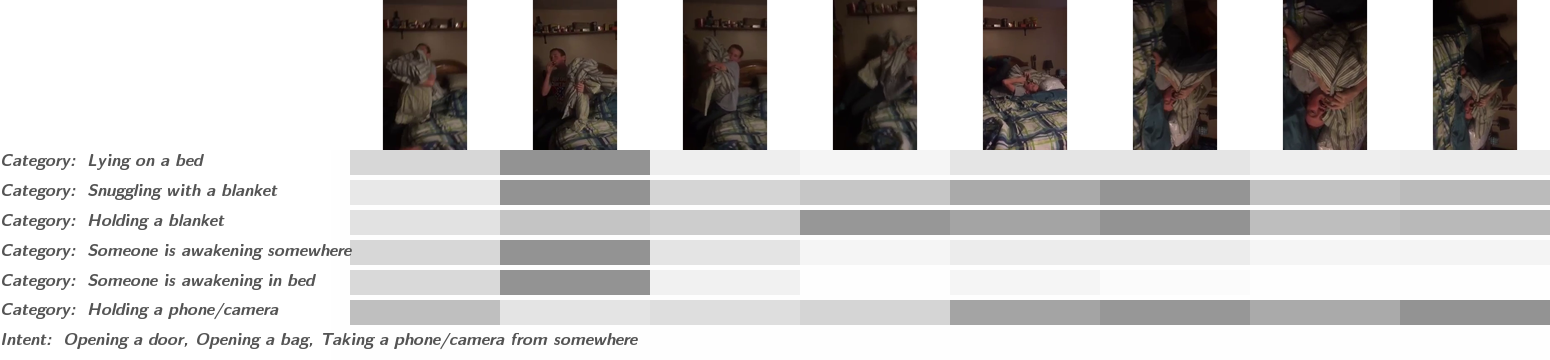}
\includegraphics[trim=0 30 0 0,clip,width=1.0\linewidth]{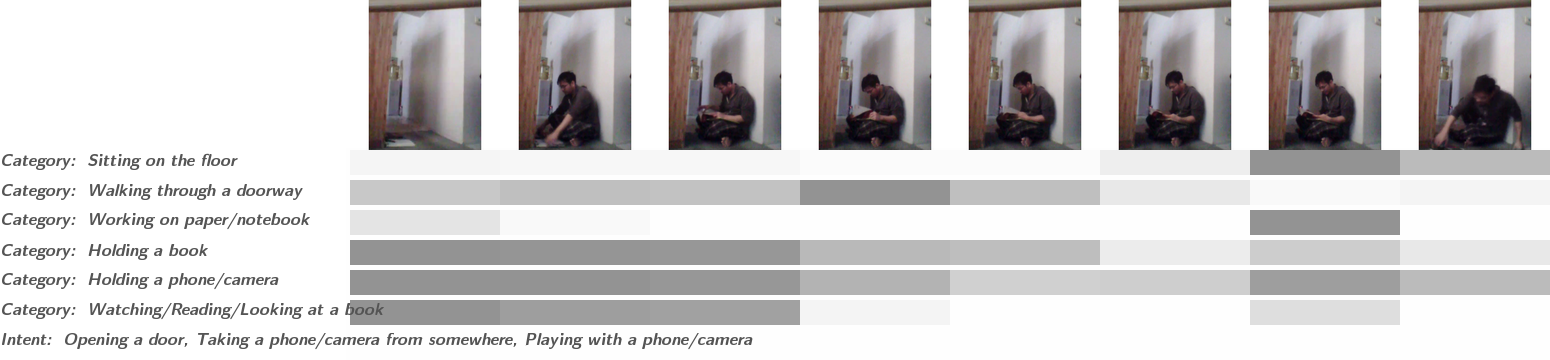}
\caption{Visualizations of the model predictions for the 6 most confident categories. Darker colors indicate higher likelihood.}
\label{fig:shortlist_actions}
\end{figure}

\begin{figure}[p]
\centering
\includegraphics[width=1.0\linewidth]{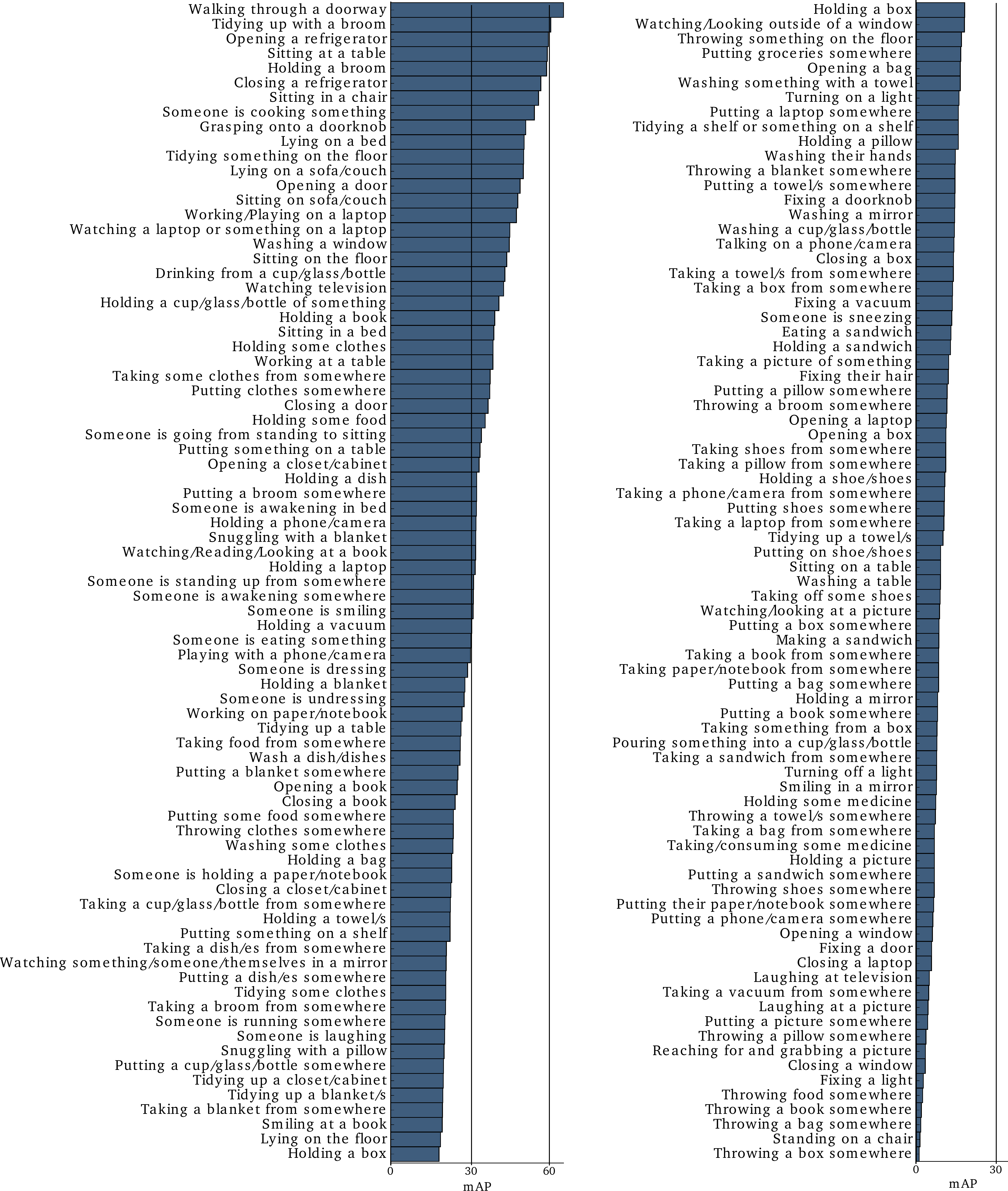}
\caption{mAP for our model for all classes, sorted by mAP. The column on the right is the continuation of the left column.}
\label{fig:mAP}
\end{figure}

Due to space constraints in the full paper, we present here additional visualizations from the model. In Figure~\ref{fig:shortlist} we present in the same way as Figure~9 (from the paper). That is, we present the 3 most confident categories, 2 most confident actions, and 1 most confident object. For example, in the first row we can see that once the light turns on in the room and the couch becomes visible the category \emph{Sitting on a sofa/couch} fires, which in turn increases the likelihood of \emph{sitting} in the next few frames. 
Furthermore, in Figure~\ref{fig:shortlist_actions} we present similar visualizations, but only the 6 most confident categories, to further understand the interplay between the activity categories. In the first row, we can see a video of a person walking towards the camera, and we can see how one after the other the model recognizes cup, phone, and sandwich, and reasons about these connected activities.
Finally, in Figure~\ref{fig:mAP} we present a breakdown of the mean average precision (mAP) by our model for each class of the dataset, sorted by the mAP of our model.

\clearpage

\end{document}